\documentclass{article}
\usepackage{amsmath}
\DeclareMathOperator*{\argmin}{argmin}  
\usepackage{graphicx}
\usepackage{epstopdf}
\usepackage{booktabs}
\usepackage{multirow}
\usepackage{subcaption}
\usepackage{float}

\usepackage[utf8]{inputenc} 
\usepackage[T1]{fontenc}    
\usepackage{hyperref}       
\usepackage{url}            
\usepackage{booktabs}       
\usepackage{amsfonts}       
\usepackage{nicefrac}       
\usepackage{microtype}      
\usepackage{xcolor}         
\usepackage{algorithm}
\usepackage{algpseudocode}
\usepackage{amsthm}

\newtheorem{theorem}{Theorem}[section]

\title{Automated Architecture Synthesis for Arbitrarily Structured Neural Networks}

\author{%
	Xinshun Liu\\
	\texttt{liuxinshun12@163.com} \\
     \and
	Yizhi Fang\\
	New York University\\
	\texttt{nooneimportant121@gmail.com} \\
     \and
	Yichao Jiang\\
	Shenzhen University of Technology\\
	\texttt{mjiang@sztu.edu.cn} \\
}

\begin{document}

	\maketitle
	
\begin{abstract}
This paper offers a new perspective on Artificial Neural Networks (ANNs) architecture. Traditional ANNs commonly use tree-like or DAG structures for simplicity, which can be preset or determined by Neural Architecture Search (NAS). Yet, these structures restrict network collaboration and capability due to the absence of horizontal and backward communication. Biological neural systems, however, feature billions of neural units with highly complex connections, allowing each biological neuron to connect with others based on specific situations. Inspired by biological systems, we propose a novel framework that learns to construct arbitrary graph structures during training and introduce the concept of Neural Modules for organizing neural units, which facilitates communication between any nodes and collaboration among modules. Unlike traditional NAS methods that rely on DAG search spaces, our framework learns from complete graphs, enabling free communication between neurons akin to biological neural networks. Furthermore, we present a method to compute these structures and a regularization technique that organizes them into multiple independent, balanced neural modules. This approach reduces overfitting and improves efficiency through parallel computing. Overall, our method allows ANNs to learn effective arbitrary structures similar to biological ones. It is adaptable to various tasks and compatible across different scenarios, with experimental results demonstrating its potential.
\end{abstract}

\section{Introduction}
This work presents a fresh viewpoint on Artificial Neural Networks (ANNs) architecture. Traditional ANNs are often arranged hierarchically in a tree-like structure or DAG, either predefined or learned through NAS methods that search within a DAG space. However, this conventional approach hinders effective communication among nodes and introduces a dramatic structural bias. In reality, existing ANN structures are optimized within a limited space, which significantly reduces the potential of ANNs and prevents them from achieving their full capabilities. Our work reimagines traditional ANNs by arguing that current connectivity approaches fail to capture the true nature of neural networks. The nodes in an asynchronous tree-like structure cannot establish connections, which impedes information transfer between neural units and leads to deficiencies. To address this limitation, we propose a method to construct a synchronous graph structure for the nodes using our introduced Neural Modules, thereby promoting collaboration among neural units. In fact, many researchers have identified the issues with the current ANN structure and have attempted to design cyclic graph structures for ANNs. However, they have not analyzed the essence of generalized cyclic structures for ANNs or proposed a framework for automatically formulating such structures as biological neural networks. These are precisely the academic attempts of this paper.

Our method provides a comprehensive search space for NAS methods. Our approach enables synchronous communication among all nodes within the structure and introduces a method for dynamically forming these structures during the learning process. These enhancements improve information transfer, thereby boosting the overall capacity of NN architectures. By fostering collaboration among nodes and automatically learning the structure, our method harnesses the collective power of neural networks, unlocking their potential in a manner akin to biological neural networks.

It is important to note that conventional tree-like neural networks are essentially a subset of our designed general graph structure. We clarify the inherent bias in current neural network architectures and provide a theoretical basis for incorporating general graph structures into neural networks. A detailed analysis of this aspect is included in the appendix. Within our framework, multiple neural units collaboratively execute precise functional implementations automatically during the learning process. Our innovation aims to bridge the gap between ANNs and more generalized structures akin to biological neural networks.

Designing this architecture presents significant challenges, as the framework introduces greater computational demands and a higher risk of overfitting. To tackle these issues, we introduce a novel regularization method. This method efficiently organizes nodes into multiple independent neural modules capable of parallel processing on modern GPU. It enables automatic node organization, thereby enhancing learning efficiency and reducing overfitting, which in turn improves overall performance.

Our learning process adapts well to larger search spaces and diverse tasks. We tested our optimization method on top-tier networks. The results show it works better than existing networks on many real-world tasks and datasets. This indicates that the connectivity our method learns is better in terms of performance and efficiency.

In summary, the key contributions of this study are:

1. We analyze the bias in existing tree-like neural network structures and provide a theoretical analysis for our proposed architectural improvements.

2. We develop a method for automatically learning to construct arbitrary graph structures for ANNs.

3. We introduce a novel regularization technique that organizes neural units into introduced concept Neural Modules, thereby enhancing structural efficiency and performance by reducing overfitting.

\section{Related Works}
To progress the existing tree-like structure for NNs. Yuan [1] recently provided a topological perspective, highlighting the benefits of dense connections offered through shortcuts in optimization [2] [3]. Furthermore, sparsity constraints have also been proven effective in optimizing learned structures across various applications [2] [4] [5] [6] [7]. In their approach, the structure of NNs is organized as a DAG, whereas we organize it as a more general graph structure. 

Furthermore, in recent years, Cyclic Structure with Forward-Forward Algorithm [40] also tried to design such structure for NN. The difference in our work can be organized as follows. First, the graph structure in [40] is predefined. On the other hand, our framework approaches the graph structure search within a complete graph space. Unlike predefined concepts, where each neural unit can connect with any other neuron, in a complete graph structure. Second, they achieve an equilibrium state through repetitive loops, but do not explain the essence of the loop or when the loop stops. In contrast, we analyze the essence of the equilibrium state exactly. Third, they do not analyze the size of cyclic graphs for the model or how to control it, which is crucial for the efficiency of the whole framework. For our framework, we proposed  NM regularization to control the complexity of the structure, thereby enhancing our model's performance in terms of both effectiveness and efficiency. The details of the analysis can also be found in the appendix.

The fixed point of the implicitly hidden layer can also serve as a solution [8] [9] [5] [10] [11] [12], as demonstrated in the subsequent works [13] [14] [15]. Departing from the infinite structure of implicitly hidden layers [5] [16] [17] [10] [18], we organize it into a general graph structure. Compared with implicitly hidden layers, our method can improve the efficiency by parallel computing as well as the performance by reducing overfitting. 

Our process also involves compressing NNs. In recent years, various algorithms have been developed, including quantization [19] [20] [21], low-rank approximation [22] [23] [24], knowledge distillation [25] [26], and network pruning [27], etc. The network pruning method we use in this work is weight pruning, which aims to eliminate weak connections. We aim to improve weight pruning for our framework using a method similar to [28], which evaluates the gradient at the pruned model and applies parameter updates to the dense model. In our framework, this process coordinates with an elegant regularization to automatically allocate Neural Modules. The process can be explained exactly in the algorithm in the appendix.

Other structure like OptNet integrates optimization quadratic problems for nodes within the same layer [29] [30]. However, this approach introduces additional bias. 

Graph Neural Networks (GNNs) specially address the needs of geometric deep learning [31] [32] [33] [34]. GNNs adapt their structure to the input graph, capturing complex dependencies [35] [34] [36] [32]. GNN primarily deals with graphs as inpput data.

The flexibility of this graph structure has also been investigated in studies similar to those on Reservoir Computing [37] [38] [39]. These studies have utilized a recurrent neural network framework where neuron connections are established randomly, where the weights remain static post-initialization. In contrast, our Neural Module framework allows for the adaptive learning of both weights and network structure during processing. 

\section{Methodology}
\subsection{The Mathematical Formalization of The Model}
Let $N^0$ be the input values fed into the input layer. Let $N^m$ be the nodes for the last layer, and they feed into the output values. 
In our work, we organize the intermediate structure as the complete graph. The model is denoted by $NMs$, $NMs =\{N^0, E^1, \mathcal{G}, E^m, N^m\}$, where $\mathcal{G} = \{E,N\}$ and $n_i \in N$ is the $i$th node in $\mathcal{G}$, $e_{ij} \in E$ is the edge from $n_i$ to $n_j$. Let the number of nodes in $N$ be $p$, the number of nodes in $N^0$ be $|N^0|$, and the number of nodes in $N^m$ be $|N^m|$. 

\subsection{Model Structure}
In our framework, nodes are initially computed based on their input nodes, which solely distribute features. Additionally, each node is influenced by other nodes in the complete graph $\mathcal{G}$, creating mutual influence.

Here, we introduce the concept of \textbf{Neural Modules (NM): let the parameter $\gamma$ be the $k$th largest absolute value of the weights in $\mathcal{G}$, a Neural Module is defined as a Strongly Connected Component of $\mathcal{G}$ and the absolute value of the weight for any edge in it exceeds the predefined threshold $\gamma$}. 

In accordance with the definition of NM, our structure is constructed as follows: all intermediate nodes are organized into a general graph $\widetilde{\mathcal{G}}$ derived from a complete graph $\mathcal{G}$. Within this configuration, each node is influenced by all other nodes in the graph through the learning process. Nodes in the general graph $\widetilde{\mathcal{G}}$ are connected via directed edges with learnable weights, allowing for dynamic information flow and feature transformation between nodes. This mechanism permits each node to not only process its own input but also to integrate information from all other nodes, leading to a more comprehensive and robust representation of the data.

In our framework, the structure emerges from searching within a complete graph set and selecting the top $k$ edges using pruning techniques in each iteration. Unlike conventional NAS methods that search within a DAG, our approach can generate arbitrary graph structures, thereby significantly reducing structural bias. The detailed process is provided in the algorithm part in the appendix.

In the upcoming section, we elaborate on the process of calculating node values in the graph $\widetilde{\mathcal{G}}$.

\subsection{Forward Process}
First, we discuss the case of the value for each node in the graph. Here in this paper, the value for each node is represented as $x$ and the value for each edge is represented as $w$ with the same corresponding edge index. As introduced in the previous section, these values depend on the nodes in $N^0$ as well as other nodes in $\widetilde{\mathcal{G}}$. Hence, we need a synchronization method to address this. We treat the problem as a system of multivariate equations. For the value of the nodes in $\widetilde{\mathcal{G}}$, we have the following:
	
\begin{equation}
	\begin{cases}
		w_{11} + \sum\limits_{j \neq 1} f(x_{j}) \cdot w_{j1} + \sum\limits_{j = 1}^{|N^0|} x^0_j \cdot w^1_{j,1} = x_{1} \\
		w_{22} + \sum\limits_{j \neq 2} f(x_{j}) \cdot w_{j2} + \sum\limits_{j = 1}^{|N^0|} x^0_j \cdot w^1_{j,2} = x_{2} \\
		... \\
		w_{pp} + \sum\limits_{j \neq p} f(x_{j}) \cdot w_{jp} + \sum\limits_{j = 1}^{|N^0|} x^0_j \cdot w^1_{jp} = x_{p} \\
	\end{cases}
\end{equation}
	
In the above equations, $w_{11}$, $w_{22}$, ..., $w_{pp}$ are the values of the self-spin edges for $\widetilde{\mathcal{G}}$ and they represent the bias of the nodes. $f$ is the activation function.

 Let $W^m$ be the weights of $E^m$ and $X = \{x_1, x_2, ..., x_p\}$ be the values of the nodes in $\widetilde{\mathcal{G}}$ . Then, the output values $\widetilde{Y} = X^m$ can be inferred by $X^m = g(f(X) \cdot W^{mT})$, where $g$ is the activation function for output.

Existing numerical methods, such as the Newton-Raphson method [41], can effectively solve the above equations. In real-world applications, besides Newton's methods, the efficiency can be optimized by iterative methods, ‌dichotomy, or ‌secant methods, if the complexity of the Coefficient Matrix is controlled well. Note that each variable is processed by an activation function $f$, making the transformation nonlinear. If we consider $\widetilde{\mathcal{G}}$ as a Neural Module, we can introduce the following theorem. 

\begin{theorem}[Universal Neural Module Approximation Theorem]
Let $\mathcal{F}$ be an implicit function defined on a compact set, capable of being transformed into a continuous explicit function for all the variables. In such a case, there exists a Neural Module that can effectively approximate the function $\mathcal{F}$.
\end{theorem} 

\subsection{Backward Process}
To determine the gradient of the nodes in the neural module, we consider the gradient of the output as the last layer $\nabla X^m = \nabla Y$. The same as the Forward Process, the nodes' gradients also interact with each other. Note that, each node has been processed by the activation functions. Then, we treat the gradient of the nodes in the graph as variables within the following system of equations. 
\begin{equation}
	\begin{cases}
		\sum\limits_{j \neq 1} \nabla x_{j} \cdot f'(x_{j}) \cdot w_{1j} + \sum\limits_{j = 1}^{|N^m|} \nabla x^m_j \cdot g'(x^m_{j}) \cdot w^m_{1j} = \nabla x_{1} \\
		\sum\limits_{j \neq 2} \nabla x_{j} \cdot f'(x_{j}) \cdot w_{2j} + \sum\limits_{j = 1}^{|N^m|} \nabla x^m_j \cdot g'(x^m_{j}) \cdot w^m_{2j} = \nabla x_{2} \\
		... \\
		\sum\limits_{j \neq p} \nabla x_{j} \cdot f'(x_{j}) \cdot w_{pj} + \sum\limits_{j = 1}^{|N^m|} \nabla x^m_j \cdot g'(x^m_{j}) \cdot w^m_{pj} = \nabla x_{p} \\
	\end{cases}
\end{equation}

Finally, we can compute the gradient of the edges of our NM. First, for the gradient of the edges in the complete graph $\mathcal{G}$, according to the system of equations, we need to consider the gradient of each node. For any $j$th node in the graph, the weights of the incoming arc are represented by the $j$th($1 \leq j \leq p$) column of its adjacency matrix. For convenience, we introduce the following operator: \begin{equation} \mathcal{H}_j=[f(x_1),...,f(x_{j-1}),1,f(x_{j+1}),...,f(x_p)] ~,\end{equation} This operator is derived from the system of equations in the forward process. Then, by the gradient of the $j$th node, its corresponding gradient for $W^T_{:j},1 \leq j \leq p$ in $\widetilde{\mathcal{G}}$ can be formulated as follows: \begin{equation} \nabla W^T_{:j}= \nabla x_j \cdot f'(x_{j}) \cdot \mathcal{H}_j ~.\end{equation}

Second, for the gradient for the edges in $E^m$, according to the process, \begin{equation} \nabla W^m= \nabla X^{mT }\cdot g'(X^{mT}) \cdot f(X) ~.\end{equation}
Third, for the gradient for the edges in $E^1$, according to the process, \begin{equation} \nabla W^1= \nabla X^{T } \cdot f'(X^T) \cdot X^0 ~.\end{equation}

\subsection{Neural Module Optimization}
In this section, we revisit the previously introduced NM regularization. For a general graph $\widetilde{\mathcal{G}}$, we first normalize its adjacency matrix to obtain $\widetilde{W}$. Subsequently, we define its distance matrix $D$, where each element $d_{ij}$  is defined as follows: \begin{equation} d_{ij} = \frac{e^{-\widetilde{w}_{ij}}}{\sum {e^{-\widetilde{w}_{ij}}}}. \end{equation}

Second, NM regularization accounts for the number of nodes in the neural module. We introduce the operator $ \mathcal{Z}=[z_1,z_2,...,z_p] $, where each element $z_i, 1 \leq i \leq p$ in $\mathcal{Z}$, $z_i$  represents the number of nodes in the neural module corresponding to node $n_i$. 

Third, based on the inverse proportionality to distance law in two-dimensional graph space, we introduce the repulsion matrix $R$ for $\widetilde{\mathcal{G}}$. Each element $r_{ij} \in R$ is defined as \begin{equation} r_{ij} = \frac{z_i*z_j}{d_{ij}}\end{equation}

Let $\alpha$ be the regularization parameter. Through NM regularization, the repulsion matrix $R$ adaptively adjusts $\alpha$ in each iteration. This process facilitates the automatic organization of the graph into balanced, properly sized subgraphs, forming rational neural modules that effectively utilize neuron units. For the $i$th node in $\widetilde{\mathcal{G}}$, $1 \leq i \leq p$,  our NM regularization is formulated as: \begin{equation} J_{NM}(x_{:i}) = J(x_{:i}) + \alpha \sum_{j=1}^{p}{ r_{ji} \cdot w_{ji}^2}, \end{equation} where $J$ represents the objective function.

During the backward propagation process, the weight for each edge is adjusted as follows:

\begin{theorem}
For NM regularization, in every iteration and with learning rate $\eta$, parameter $w_{ij}$ upgrades as follow: \begin{equation} w_{ij} \gets w_{ij}(1-\eta \alpha r_{ij}) - \eta\frac{\partial J}{\partial w_{ij}}\end{equation}
\end{theorem} 

Thus, when $r_{ij}$ takes a higher value, $w_{ij}$ is more likely to tend towards zero.

To simplify the analysis and enhance understanding of NM regularization, we propose a theorem by omitting the activation function and instead analyzing its effect under linear regression.

\begin{theorem}
Under linear regression with $y = X w_{:i} + \epsilon$, with $\epsilon \sim \mathcal{N}(0,\delta^2I)$ and $w_{:i} \sim \mathcal{N}(0,\tau^2 diag(r_{:i})^{-1})$ under NM regularization, the weight distribution is given by: \begin{equation} w_{:i} \sim \mathcal{N}((X^T X + \lambda diag(r_{:i}))^{-1}X^Ty,\delta^2(X^TX + \lambda diag(r_{:i}))^{-1}), \end{equation} where $\lambda = \frac{\delta^2}{\tau^2}.$
\end{theorem} 

From the theorem above, a larger repulsion term in $R$ makes the expectation of the weight closer to zero and reduces its variance.

In NM regularization, for each element $r_{ij}$ in the repulsion matrix $R$, a larger $z_i$ or $z_j$ increases its value, resulting in a stronger repulsive force. Similarly, smaller elements in the distance matrix $D$ have the same effect. The effects can be organized into the following two aspects.

\textbf{Between Neural Modules}: NM regularization prevents the formation of overly large neural modules, as previously analyzed.

\textbf{Within Neural Modules}: NM regularization helps prevent the adjacency matrix from becoming overly complex. This is important because overly complex adjacency matrices can lead to instability in solutions within neural modules, which is often caused by coefficient matrices with high condition numbers.

To provide convergence guarantees for NM regularization, we establish theoretical bounds based on the following assumptions: The training objective is smooth, satisfying $\|\nabla f(w) - \nabla f(v)\| \leq L \|w-v\|,\forall w,v \in \mathbb{R}^p,$ for some constant $L > 0$. The stochastic gradients are bounded, with $\mathbb{E}\|\nabla w\|^2 \leq G^2$. Under these assumptions, the convergence of NM regularization is presented in the following theorem.

\begin{theorem}
Let the learning rate be $\gamma = \frac{c}{\sqrt{T}}$, where $c = \sqrt{\frac{f(w_0)-f(w_*)}{LG^2}} and $ $T$ is the number of iterations. For a pruned model selected according to the definition of Neural Module, the following inequality holds in expectation over the selected edges $u$: \begin{equation} \mathbb{E}\|\nabla u\|^2 = \mathcal {O}(\sqrt{\frac{L(f(w_0)-f(w_*))}{T}}G + L^2\|w_0 - \tilde{w_0}\|^2, \end{equation}
\end{theorem}

\subsection{Algorithm}

In this section, we present the algorithm for our framework. First, initialize the adjacency matrix of the complete graph $\mathcal{G}$ as $\mathcal{K}$. Then, approximate $\mathcal{K}$ to the coefficient matrix $\mathcal{C}$ by parameter $\gamma$, eliminating all edges whose weights have absolute values lower than $\gamma$.

The computation of the graph is based on a method akin to topological sorting. For each neural module $NM_i$, $InArc(NM_i, \mathcal{C})$ is the number of edges $e_{js}$  in the approximated graph that connect from node $n_j \notin NM_i$ to node $n_s \in NM_i$. Conversely, $OutArc(NM_i, \mathcal{C})$ represents the number of edges $e_{js}$ in the approximated graph that connect from node $n_j \in NM_i$ to node $n_s \notin NM_i$. Here, |NM| denotes the number of nodes within the neural module.

In our algorithm, Tarjan's algorithm [42] is employed to identify strongly connected components, which subsequently serve as the foundation for forming neural modules.

The algorithm is founded on a generalized form of topological sorting. Each iteration consists of two main processes. The first process involves propagating according to the traditional neural network (NN) process. The second process aims to handle Neural Modules as a system of equations. During each iteration, the algorithm updates by removing related edges. Note that the diagonal elements represent biases that have already been accounted for in the first process and should not influence the second process. In summary, our framework is detailed in Algorithm 1, Algorithm 2, and Algorithm 3, which are provided in the appendix.

For Algorithm 1, the complexity analysis of the main part is as follows. The complexity of Tarjan's algorithm is $ \mathcal{O}(|N|+|E|)$, and the complexity of checking InArc or OutArc is $\mathcal{O}(|E|)$. The complexity of solving the system of equations can be optimized to $\mathcal{O}(|NM|^2)$, where the complexity for each neural module (NM) has been optimized through NM regularization. Therefore, in the worst-case scenario, where the structure is organized as a list that prevents parallel computing, the overall complexity of our propagation algorithm is $\mathcal{O}(|N|+|E|+q*|NM|^2)$ where $q$ is the number of neural modules. If the topology supports parallel computing, the overall complexity can be optimized to $\mathcal{O}(|N|+|E|+max(|NM|^2)$.

\section{Experiments}
In this section, we present experiments conducted with our Neural Modules, comparing their performance with traditional NN methods and some state-of-the-art models, including implicit hidden layers (DEQ), a topological perspective treating NN as a DAG, as well as recently introduced OPTNET. The results are tabulated for five real datasets, all of which are available in the UCI dataset.

The five datasets in our experiments are for different Real-life scenarios. The first dataset comprises codon usage frequencies in genomic coding DNA from a diverse sample of organisms obtained from different taxa in the CUTG database. The second dataset involves smartphone-based recognition of human activities and postural transitions, performing various activities. The third dataset Facebook Large Page-Page Network contains a webgraph. This webgraph is a page-page graph of verified Facebook sites. Nodes represent official Facebook pages, while the links are mutual likes between sites. Node features are extracted from the site descriptions that the page owners created to summarize the purpose of the site. The fourth dataset Daily and Sports Activities dataset, comprises motion sensor data of 19 daily and sports activities, each performed by 8 subjects in their own style for 5 minutes. The fifth dataset includes measurements from 16 chemical sensors exposed to six different gases at various concentration levels. All these tasks represent classification problems, and we evaluate performance based on the error of each algorithm. Due to space constraints, experimental results for the fifth dataset are provided in the appendix.
	
\subsection{Performance of Neural Modules with Layer Scale}
Since the major compared methods, such as DEQ or OPTNET, operate at the layer scale, these state-of-the-art methods primarily focus on individual layers within neural networks. At this scale, the number of nodes typically does not exceed several hundred. In this section, we compare our framework with these state-of-the-art methods in the context of neural networks with fewer than 300 nodes. By analyzing the results, we can substantiate our claims of having significant advantages at this node scale.

We assessed the effectiveness of our NMs and other methods across various node complexities. This evaluation allows us to understand how the NMs perform with different levels of complexity. The nodes were initially organized using NN, DEQ, DAG as well as OPTNET, and their percentage of error was observed. Our experiments involved comparing the performance of NMs, examining their performance with frequently-used $L1$ regularization and the proposed NM regularization, as introduced earlier. The results, as presented in Table 1, reveal that our novel structure consistently achieves superior results in most cases. Additionally, the performance of NM regularization surpasses that of the benchmark. With the optimal performance achieved with a suitable number of nodes, our NM regularization is shown to perform the best across all the datasets. 

From Table 1, we can get the conclusion that NN consistently exhibits improved results when nodes are organized into neural modules. The performance of our NMs can be further enhanced through regularization, as previously explained. Notably, even without regulation, our NMs outperform the traditional structure. After implementing NM regularization, our approach demonstrates enhanced performance. The error bars for these data are lower than 0.007.  

\begin{table}[!htbp]
	\centering
     \scriptsize
	\caption{The performance of algorithms}
	\begin{tabular}{cccccc|ccccc} 
		\toprule
             &      &     & Condon     &      &     &     &     & Postual     &      &     \\
	Methods &  60  & 80  & 100  & 200  & 300 & 60  & 80  & 100  & 200  & 300    \\
	\midrule
	NN      &  0.1917  & 0.1854 & 0.1964 & 0.1702  & 0.1839  & 0.1099 & 0.1008 & 0.1069 & 0.0811 & 0.1245 \\
	DEQ     &  0.1839 & 0.2136 & 0.1980  & 0.1714 & 0.2211  & 0.2755 & 0.0882 & 0.0889 & 0.0859  & 0.0958 \\
	DAG     &  0.2152 & 0.2027 & 0.2050 & 0.2074  & 0.1987 & 0.1130 & 0.1184 & 0.0937 & 0.1025  & 0.1639 \\
	OPTNET  &  0.1706 & 0.2152 & 0.2034 & 0.1745  & 0.1901 & \textbf{0.1055} & 0.0940 & 0.1164 & 0.0994  & 0.1622 \\
	NMs     &  0.1498 & 0.1487 & 0.1465 & 0.1387 & 0.1398   & 0.2678 & 0.0876 & 0.0912 & 0.0901 & 0.0956 \\
	NMs\&L1 &  \textbf{0.1410} & 0.1422 & 0.1395 & 0.1366  & 0.1345   & 0.2367 & \textbf{0.0866} & 0.0892 & 0.0802  & 0.0888 \\
	NMs\&NM &  0.1456 & \textbf{0.1378} & \textbf{0.1315} & \textbf{0.1215} & \textbf{0.1277}  & 0.2416 & 0.0888 & \textbf{0.0672} & \textbf{0.0768} & \textbf{0.0876} \\
     \midrule
             &      &     &  Facebook    &      &     &     &     &  Activity    &      &     \\
	Methods & 60  & 80  & 100  & 200  & 300  & 60  & 80  & 100  & 200  & 300    \\
	\midrule
	NN      & 0.1459 & 0.1502 & 0.1329 & 0.1385 & 0.1514   & 0.4553 & 0.4142 & 0.4089 & 0.3160 & 0.3027 \\
	DEQ     & 0.1410 & 0.1510 & 0.1366 & 0.1310  & 0.1699  & 0.4478 & 0.3620 & 0.3444 & 0.3058  & 0.2858 \\
	DAG     &  0.1639 & 0.1554 & 0.1528 & 0.1485  & 0.1798  &  0.6478 & 0.6010 & 0.5878 & 0.5678  & 0.5510 \\
	OPTNET  &  0.1381 & 0.1615 & 0.1472 & 0.1576  & 0.1619 &  0.4273 & 0.3913 & 0.3618 & 0.3429  & 0.2758 \\
	NMs     & \textbf{0.1298} & 0.1356 & 0.1230 & 0.1326 & 0.1467    & 0.4195 & 0.3887 & 0.3367 & 0.3167 & 0.2987 \\
	NMs\&L1 & 0.1387 & 0.1345 & 0.1315 & 0.1356  & 0.1420  & \textbf{0.3989} & 0.3676 & 0.3567 & 0.3078  & 0.2867 \\
	NMs\&NM & 0.1312 & \textbf{0.1267} & \textbf{0.1212} & \textbf{0.1231} & \textbf{0.1166} & 0.4012 & \textbf{0.3566} & \textbf{0.3096} & \textbf{0.2867} & \textbf{0.2567} \\
	\bottomrule
\end{tabular}
\label{tbl:table1}
\end{table}

\subsection{Performance of Neural Modules with Network Scale}

Furthermore, for networks with a higher number of nodes, we also examine the performance of our framework on a network scale. In our experiments, parallel NM regularization also yields increasingly better results, as demonstrated in Figure 1. In our analysis, we compared methods such as DEQ, DAG, and OPTNET, and observed that when the number of nodes exceeds 300, these methods generally result in overtime. We established the performance of these models with node counts within 300 as our baseline for comparison.

\begin{figure*}[!htbp]
     \centering
     \begin{subfigure}[b]{0.4\textwidth}
        \centering
        \includegraphics[width=\linewidth]{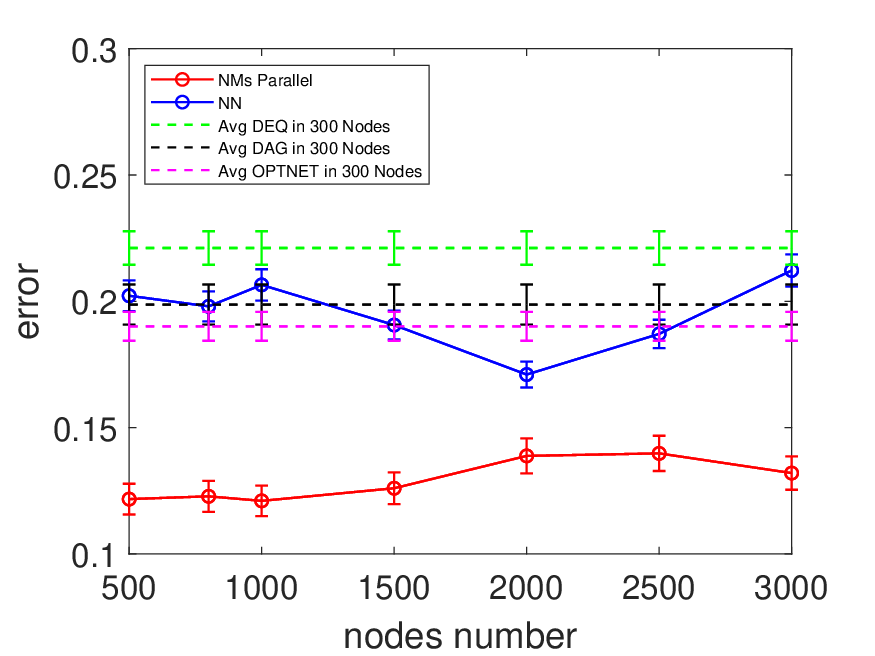} 
        \caption{Codon Usage Dataset}
        \label{Codon Usage Dataset}
     \end{subfigure}
     \begin{subfigure}[b]{0.4\textwidth}
        \centering
        \includegraphics[width=\linewidth]{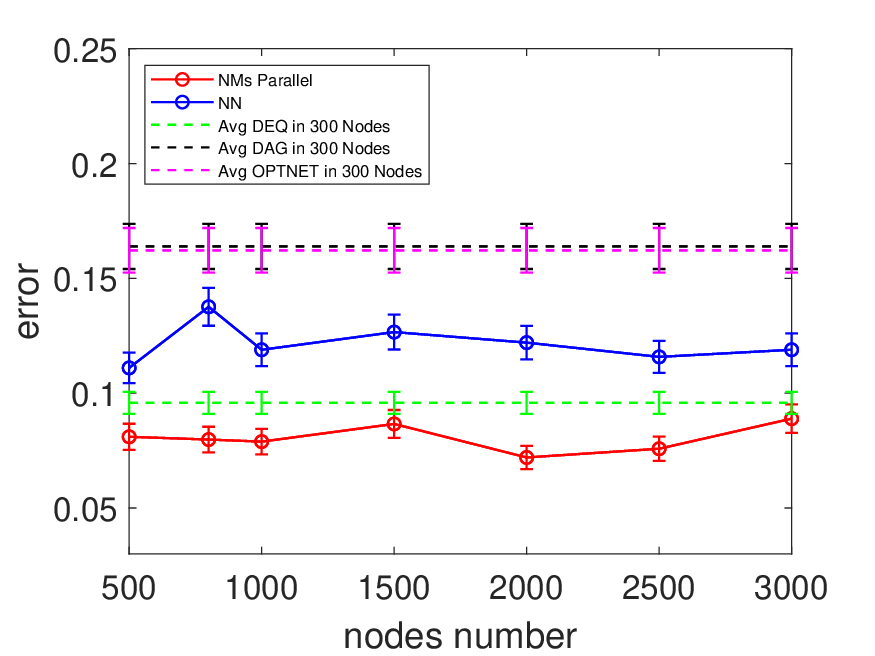} 
        \caption{Postural Transitions Dataset}
        \label{Postural Transitions Dataset}
     \end{subfigure}
     \begin{subfigure}[b]{0.4\textwidth}
        \centering
        \includegraphics[width=\linewidth]{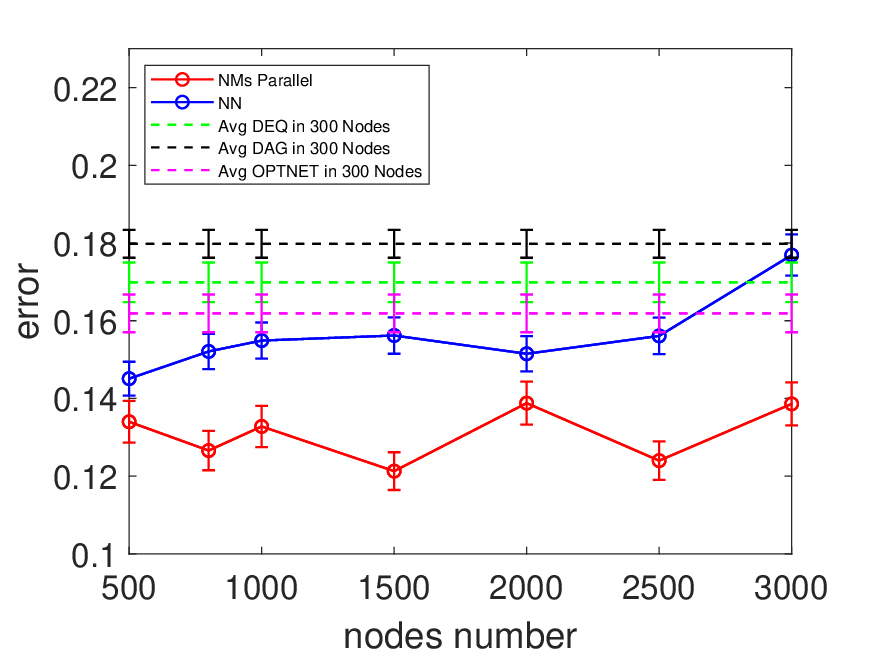} 
        \caption{Facebook Dataset}
        \label{Facebook Dataset}
     \end{subfigure}
     \begin{subfigure}[b]{0.4\textwidth}
        \centering
        \includegraphics[width=\linewidth]{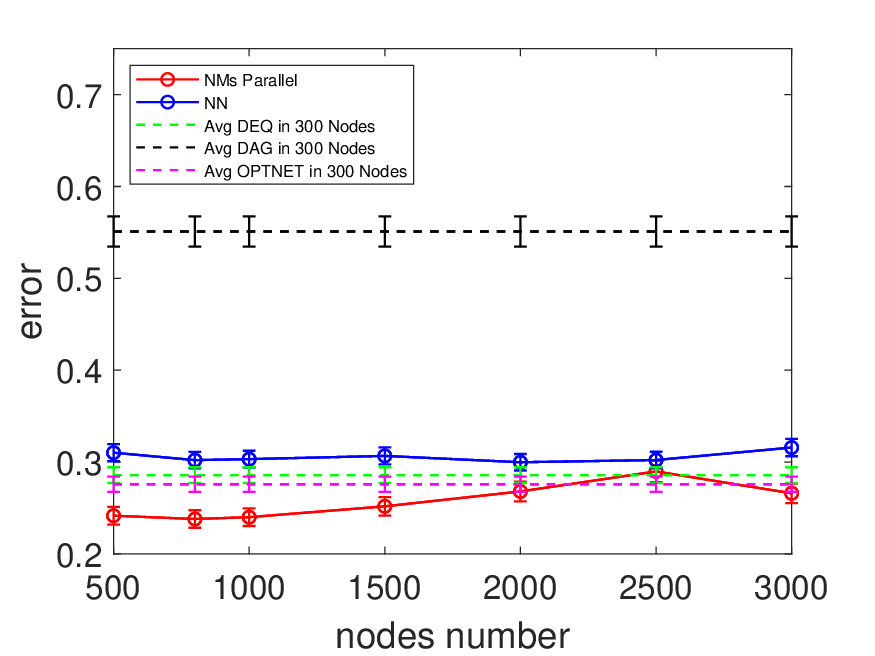} 
        \caption{Activity Dataset}
        \label{Activity Dataset}
     \end{subfigure}

      \caption{ Compared methods such as DEQ, DAG, and OPTNET generally result in overtime when the number of nodes reaches up to 3000. To establish a baseline for comparison, we calculated the performance of these three models with node counts within 300. Our findings indicate that our framework outperforms existing methods in terms of performance. Additionally, our Neural Modules (NMs) demonstrate superior efficiency. Unlike DEQ, DAG, and OPTNET, which primarily operate on a layer scale for tree-like structures, our method can be applied on a network scale to enhance the entire neural network structure. }

\end{figure*} 

Additionally, to evaluate the performance of our framework on applications with larger node numbers, we included the NN structure as a reference in our experiments. Overall, we found that our framework performs better on larger models, indicating its superior accuracy, especially when handling complex, large-scale neural network structures.

From these experiments, we believe that our Neural Module framework can be used to effectively progress existing neural networks from tree-like structures to more general graph structures.

\subsection{Efficiency of Neural Modules}
In our framework, the model can be generalized to the network scale through NM regularization, as previously introduced.

In this section, we delve into the optimization of Neural Module efficiency through the application of NM regularization. As previously mentioned, every structure considered is a subgraph of a fully connected graph, with the initial general graph acting as the search space for our model. Our NM regularization serves as a potent mechanism for structural optimization, enhancing the effectiveness and balance of neural modules. The performance of these modules, particularly for nodes numbering less than 300, was demonstrated in the preceding section, where it was shown to yield superior outcomes.

To assess the efficiency of our NM regularization, we present a comparison of the model's running time across varying complexities, represented by node counts below 300. Figure 2, Image A, mesured by seconds,illustrates that the efficiency of NM regularization significantly outperforms DEQ, especially with a larger number of nodes. This superiority is attributed to NM regularization's ability to create multiple independent neural modules, which efficiently mitigate computational complexity.

For networks with a larger number of nodes, we can leverage the parallel processing capabilities of Neural Modules to enhance the efficiency of our framework, as previously discussed. NM regularization facilitates the creation of multiple independent and well-balanced neural modules, which are inherently suited for parallel computing, particularly when transitioning to GPU-based computation. In this extension, we increase the node count from 300 to 3000 and incorporate GPU hardware acceleration to compute the algorithms more efficiently.

\begin{figure}[!htbp]
     \centering
     \begin{subfigure}[b]{0.4\textwidth}
     \includegraphics[width=\textwidth]{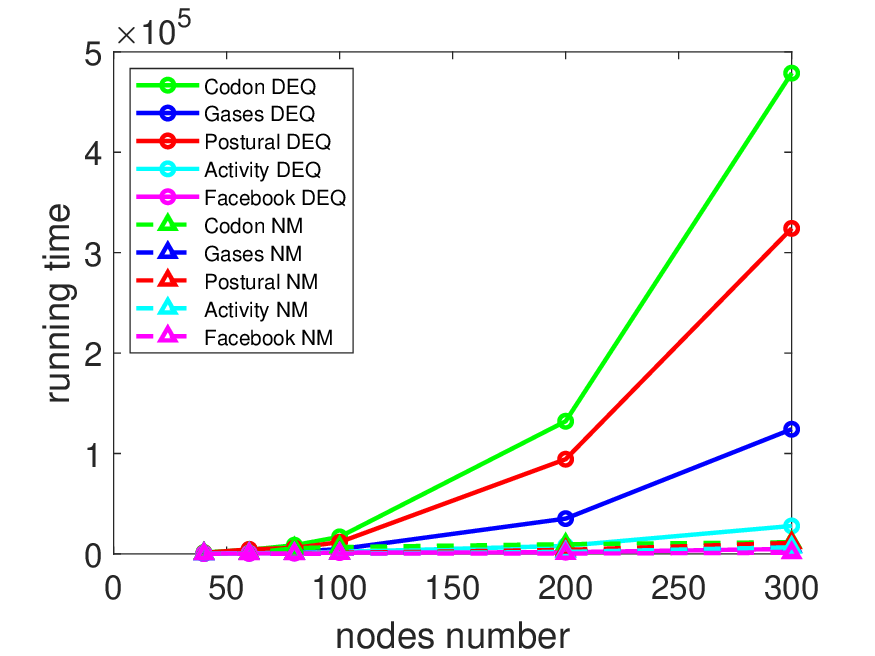}
     \caption{Nodes' number up to 300. Neural modules vs.DEQ.}
     \end{subfigure}
     \begin{subfigure}[b]{0.4\textwidth}
	\includegraphics[width=\textwidth]{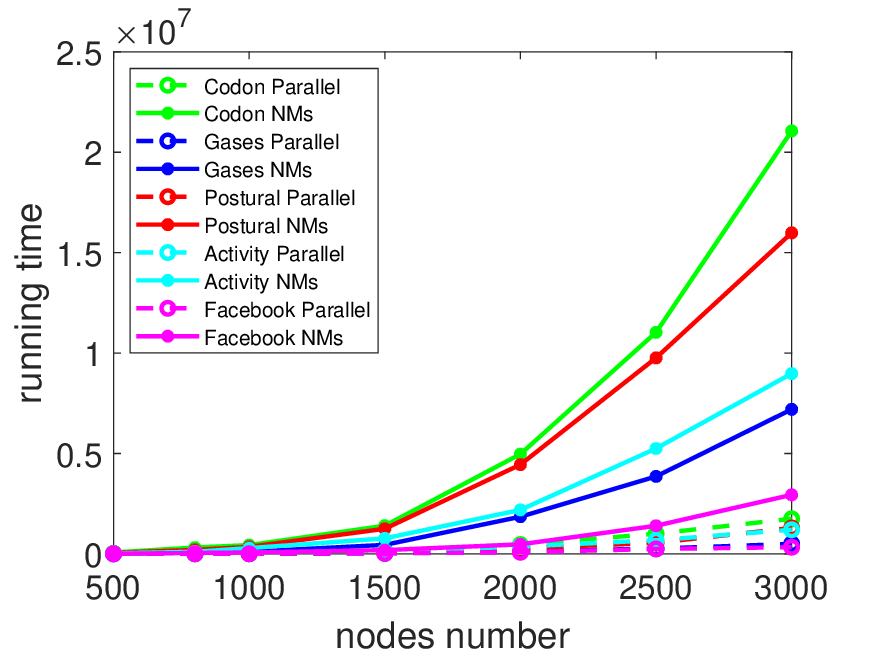}
     \caption{Nodes' number up to 3000. NM parallel vs. Neural modules.}
     \end{subfigure}
      \caption{ Neural Modules offer significant advantages over DEQ in terms of computational efficiency. Moreover, through parallel computation, NM regularization provides an opportunity to extend the applicability of this framework to larger-scale applications. By harnessing the power of parallel processing, NM regularization allows for the efficient handling of increased complexity, making it a promising approach for tackling more extensive computational challenges. }
\end{figure} 

In Figure 2, Image B, mesured by seconds, we conducted a comparison between the running time of Neural Modules operating in parallel and Neural Modules without parallelization on large realistic models. For these experiments, we utilized 16 threads. Note that the method of Neural Modules without parallel computing would result in overtime for larger nodes. Here we approximated the running time by part data here. Our results indicate that the parallel implementation of NM regularization offers a computational speedup of approximately 10 times. This demonstrates the substantial efficiency gains achievable through parallel processing in the context of NM regularization.

From these experiments, it is evident that our neural module framework can dramatically enhance the performance of Neural Networks by evolving from tree-like structures to more general graph structures. Despite the introduction of additional parameters, as analyzed in the previous sections, our framework enables the Neural Network structure to explore a nearly complete search space, effectively reducing the bias associated with the current tree-like scenario and getting much better performance. Moreover, the NM regularization and parallel computing techniques we have introduced further empower our method, allowing it to be effectively applied to larger networks. This sets our approach apart from existing state-of-the-art methods, which are often more complex and primarily focused on the layer scale of tree-like structures.

\section{Conclusion}
This study introduces a novel general graph structure designed for NNs, aiming to improve performance by facilitating significant information transfer. We address structural bias analysis for the current tree-like structure. Our model employs a synchronization method for the simultaneous calculation of node values, thereby fostering collaboration within neural modules. Additionally, we propose a novel NM regularization method that encourages the learned structure to prioritize critical connections and automatically formulate multiple independent, balanced neural structures, which would help to achieve better efficiency by calculation in parallel. This approach not only reduces the computational load associated with managing more nodes but also improves the performance by mitigating overfitting. Quantitative experimental results affirm the superiority of our proposed method over traditional structures for NNs.

\section*{References}{
\small

[1] Jing Shao Kun Yuan, Quanquan Li and Junjie Yan. 2020. Learning Connectivity
of Neural Networks from a Topological Perspective. In ECCV.

[2] Rupesh Kumar Srivastava, Klaus Greff, and Jürgen Schmidhuber. 2015. Highway
Networks. Computer Science (2015).

[3] Mark Sandler, Andrew Howard, Menglong Zhu, Andrey Zhmoginov, and
Liang Chieh Chen. 2018. MobileNetV2: Inverted Residuals and Linear Bottle necks. In 2018 IEEE/CVF Conference on Computer Vision and Pattern Recognition
(CVPR).

[4] Haoyu Chu, Shikui Wei, and Ting Liu. 2023. Learning Robust Deep Equilibrium
Models. arXiv preprint arXiv:2304.12707 (2023).

[5] K. Ahmed and L. Torresani. 2018. Mask connect Connectivity learning by gradient
descent. In ECCV.

[6] Zhang X. Ren S. He, K. and J. Sun. 2016. Deep residual learning for image
recognition.. In CVPR.

[7] Liu Z. VanDer Maaten L. Huang, G. and K.Q.. Weinberger. 2017. Densely connected convolutional networks. In CVPR.

[8] Shaojie Bai, J. Zico Kolter, and Vladlen Koltun. 2019. Deep Equilibrium Models.
In NeurIPS.

[9] Russell Tsuchida and Cheng Soon Ong. 2022. When are equilibrium networks
scoring algorithms?. In NeurIPS 2022 Workshop on Score-Based Methods.

[10] Zonghan Yang, Tianyu Pang, and Yang Liu. 2022. A Closer Look at the Adversarial Robustness of Deep Equilibrium Models. Advances in Neural Information
Processing Systems 35 (2022), 10448–10461.

[11] Howard Heaton, Samy Wu Fung, Aviv Gibali, and Wotao Yin. 2021. Feasibility based fixed point networks. Fixed Point Theory and Algorithms for Sciences and
Engineering 2021, 1 (2021), 1–19.

[12] Nicolas Zucchet and João Sacramento. 2022. Beyond backpropagation: bilevel optimization through implicit differentiation and equilibrium propagation. Neural
Computation 34, 12 (2022), 2309–2346.

[13] Shaojie Bai, Vladlen Koltun, and J Zico Kolter. 2020. Multiscale deep equilibrium
models. Advances in Neural Information Processing Systems 33 (2020), 5238–5250.

[14] Béla J Szekeres and Ferenc Izsák. 2024. On the computation of the gradient in
implicit neural networks. The Journal of Supercomputing (2024).

[15] Li P. Pang T. Yang, Z. and Y. Liu. 2023. Improving adversarial robustness of
deep equilibrium models with explicit regulations along the neural dynamics. In
ICML.

[16] Zenan Ling, Zhenyu Liao, and Robert C Qiu. 2023. On the Equivalence between
Implicit and Explicit Neural Networks: A High-dimensional Viewpoint. arXiv
preprint arXiv:2308.16425 (2023).

[17] Shutong Ding, Tianyu Cui, Jingya Wang, and Ye Shi. 2023. Two Sides of The
Same Coin: Bridging Deep Equilibrium Models and Neural ODEs via Homotopy
Continuation. arXiv preprint arXiv:2310.09583 (2023).

[18] iaming Liu, Xiaojian Xu, Weijie Gan, Ulugbek Kamilov, et al. 2022. Online
deep equilibrium learning for regularization by denoising. Advances in Neural
Information Processing Systems 35 (2022), 25363–25376.

[19] Yixing Xu Chunjing Xu Enhua Wu Kai Han, Yunhe Wang and Chang Xu. 2020.
Training binary neural networks through learning with noisy supervision. In
ICML.

[20] Chunjing Xu Mingzhu Shen, Kai Han and Yunhe Wang. 2019. Searching for accurate binary neural architectures. In IEEE International Conference on Computer
Vision Workshops. 0–0, 2019.

[21] Xuanyi Dong Yanwei Fu Yang He, Guoliang Kang and Yi Yang. 2018. Soft filter
pruning for accelerating deep convolutional neural networks. In IJCAI.

[22] Chong Li and CJ Richard Shi. 2018. Constrained optimization based low-rank
approximation of deep neural networks. In ECCV

[23] Chuanjian Liu Hanting Chen Chunjing Xu Boxin Shi Chao Xu Zhaohui Yang,
Yunhe Wang and Chang Xu. 2019. Legonet: Efficient convolutional neural
networks with lego filters. In ICML.

[24] Xinchao Wang Xiyu Yu, Tongliang Liu and Dacheng Tao. 2017. On compressing
deep models by low rank and sparse decomposition. In CVPR.

[25] Shan You Shumin Kong, Tianyu Guo and Chang Xu. 2020. Learning student
networks with few data. In AAAI.

[26] Chao Xu Shan You, Chang Xu and Dacheng Tao. 2018. Learning with single teacher multi-student. In AAAI.

[27] Stephen Tyree Iuri Frosio Pavlo Molchanov, Arun Mallya and Jan Kautz. 2019.
Importance estimation for neural network pruning. In CVPR.

[28] Luis Barba Daniil Dmitriev Martin Jaggi Tao Lin, Sebastian U. Stich. 2020. Dynamic Model Pruning with Feedback. In ICLR.

[29] Brandon Amos and J Zico Kolter. 2017. Optnet: Differentiable optimization as a
layer in neural networks. In International Conference on Machine Learning. PMLR,
136–145.

[30] Yan J. Luo C. Chen L. Lin Q. Yan, K. and D. Zhang. 2021. A surrogate objective
framework for prediction+optimization with soft constraints.

[31] M. Gori, G. Monfardini, and F. Scarselli. 2005. A new model for learning in graph
domains. In IEEE International Joint Conference on Neural Networks.

[32] Wenqi Fan, Yao Ma, Qing Li, Yuan He, Eric Zhao, Jiliang Tang, and Dawei Yin.
2019. Graph neural networks for social recommendation. (2019), 417–426.

[33] Franco Scarselli, Marco Gori, Ah Chung Tsoi, Markus Hagenbuchner, and
Gabriele Monfardini. 2008. The graph neural network model. IEEE transactions on neural networks 20, 1 (2008), 61–80.

[34] Sergi Abadal, Akshay Jain, Robert Guirado, Jorge López-Alonso, and Eduard
Alarcón. 2021. Computing graph neural networks: A survey from algorithms to
accelerators. ACM Computing Surveys (CSUR) 54, 9 (2021), 1–38.

[35] Sweah Liang Yong, Markus Hagenbuchner, Ah Chung Tsoi, Franco Scarselli, and
Marco Gori. 2007. Document mining using graph neural network. In Comparative
Evaluation of XML Information Retrieval Systems: 5th International Workshop of
the Initiative for the Evaluation of XML Retrieval, INEX 2006, Dagstuhl Castle,
Germany, December 17-20, 2006, Revised and Selected Papers 5. Springer, 458–472.

[36] Alex Fout, Jonathon Byrd, Basir Shariat, and Asa Ben-Hur. 2017. Protein interface
prediction using graph convolutional networks. In Neural Information Processing
Systems.

[37] Mantas Lukoševicius and Herbert Jaeger. 2009. Reservoir computing approaches
to recurrent neural network training. In Computer Science Review, 3(3):127–149,
2009.

[38] David Verstraeten Benjamin Schrauwen and Jan Van Campenhout. 2007. An
overview of reservoir computing: theory, applications and implementations. In
In Proceedings of the 15th european symposium on artificial neural networks. p.
471-482 2007, pages 471–482, 2007.

[39] Danilo Vasconcellos Vargas Heng Zhang. 2023. A Survey on Reservoir Computing
and its Interdisciplinary Applications Beyond Traditional Machine Learning. In
arXiv:2307.15092.

[40] Yang L, Zhang H, Zhang W, et al. Beyond Directed Acyclic Computation Graph with Cyclic Neural Network[J]. 2024

[41] Shashank Gawade. 2024. The Newton-Raphson Method: A Detailed Analysis. In International Journal for Research in Applied Science and Engineering Technology 12(11):729-734.

[42] Robert E. Tarjan. 1972. Depth-First Search and Linear Graph Algorithms. In SIAM Journal on Computing 1 (1972), 146–160.

[43] G.. Cybenko. 1989. Approximation by superpositions of a sigmoidal function. In Mathematics of control, signals and systems.

[44] Stinchcombe M. Hornik, K. and H. White. 1990. Universal approximation of an unknown mapping and its derivatives using multilayer feedforward networks. In Neural networks.

[45] Geoffrey Hinton, The Forward-Forward Algorithm: Some Preliminary Investigations. NeurIPS 2022

}


\appendix

\section{The Bias of the Traditional Tree-like Structure}
In this section, we argue that the existing tree-like neural network (NN) structure is essentially a special case of our proposed framework for solving a system of equations. For a tree-like structure with $m$ levels, let the intermediate values at the $i$th level be denoted by $X^i$, the input values be $X^0$ and the weights associated with the values at the $i$th level be represented by $W^i$. For the sake of brevity, the biases are omitted here. The asynchronous process for the current tree-like structure can be formalized by a system of equations in our framework as follows,

\begin{equation}
	\begin{cases}
		X^0 \cdot W^{1T} = X^{1} \\
		X^1 \cdot W^{2T} = X^{2}  \\
		... \\
		X^{m-1} \cdot W^{mT} = X^{m}.  \\
	\end{cases}
\end{equation}

Formating the inputs and the values of all nodes in the Neural Network as $\mathcal{X}$, the system of equations can be simplified to $\mathcal{X} \cdot \mathcal{C}^T = 0$, where $\mathcal{X}$ is a variable vector $\mathcal{X} = (X^{0}, X^{1}, X^{2}, ..., X^{m})$. $E$ is the identity matrix for each level.

Furthermore, the coefficient $\mathcal{C}$ for the tree-like structure can be represented by

\begin{figure}[h]
\[
\begin{pmatrix}
  W^1 & -E^1 & 0 & 0 & ... & 0 & 0  \\
  0 & W^2 & -E^2 & 0 & ... & 0 & 0  \\
  0 & 0 & W^3 & -E^3 & ... & 0 & 0  \\
  ... ...  \\
  0 & 0 & 0 & 0 & ... & W^{m} & E^{m}  \\
\end{pmatrix}
\]
\caption{The Coefficient $\mathcal{C}$ for tree-like structure}
\end{figure}

From the system of equations, we can see that the traditional NN structure represents a special case of our system of equations that can solved asynchronously. In the tree-like structure, the coefﬁcient matrix of the equations is composed of the parameter matrices for each level, and these coefficients are positioned near the diagonal. 

The tree-like structure in our framework is indeed more constrained for each node within the structure. This implies that for any nodes in a tree structure, if they are solely dependent on the neurons in the preceding layer, their capacity for complex function approximation is severely limited. 

Some existing work has attempted to generalize the structure to a Directed Acyclic Graph (DAG), such as in the case of ResNet. In these architectures, additional weights are introduced to the Lower Triangular of the coefficient matrix $\mathcal{C}$. Let the additional weights for each level be formulated as $V$. The matrix $\mathcal{C}$ can be depicted as follow:

\begin{figure}[h]
\[
\begin{pmatrix}
  W^1 & -E^1 & 0 & 0 & ... & 0 & 0  \\
  V^1 & W^2 & -E^2 & 0 & ... & 0 & 0  \\
  V^1 & V^2 & W^3 & -E^3 & ... & 0 & 0  \\
  ... ...  \\
  V^1 & V^2 & V^3 & V^4 & ... & W^{m} & E^{m}  \\
\end{pmatrix}
\]
\caption{The Coefficient $\mathcal{C}$ for DAG or ResNet}
\end{figure}

 This modification allows for a more flexible and expressive model, which can better approximate complex functions and handle larger datasets, aligning with the Approximation Theorem's [43][44] assertion that a single hidden layer with a sufficient number of neurons can approximate any continuous function on a compact subset of the real numbers. These methods are designed to enhance the expressive capability of each neural unit in this manner.

The Lower Triangular Coefficient matrix indeed reaches the limits of asynchronous structures. However, the potential of the Upper Triangular matrix remains largely untapped. In our work, we extend the asynchronous structure to a synchronous one, breaking through the limitations of the Lower Triangular and generalizing the coefficient matrix $\mathcal{C}$. Within our Neural Module framework, the $\mathcal{C}$ is depicted in Figure 5, showcasing a more comprehensive and interconnected structure that allows for greater flexibility and performance. Our framework generalizes the coefﬁcient to form a full matrix across all levels, indicating a transition from a tree-based structure to a general graph structure.

\section{The Rationale of Introducing Neural Modules}
For the traditional tree-like structure, the asynchronous forward and backward process can also be considered to solve a system of equations. In our framework, the NN structure is a general graph with $p$ nodes.

For our model. the diagonal of the adjacency matrix of the Neural Network can be formulated as the bias for each node. Assign the input and bias-related values to the right side. Allocate the values associated with each node to the left side. After each neural unit in NN is processed by input $X^0$. Consequently, the coefficient matrix $\mathcal{C}$ becomes the matrix in Figure 5.

\begin{figure}[h]
\[ 
\begin{pmatrix}
  -1 & w_{2,1} & w_{3,1} & w_{4,1} & ... & w_{p-1,1} & w_{p,1}  \\
  w_{1,2} & -1 & w_{3,2} & w_{4,2} & ... & w_{p-1,2} & w_{p,2}  \\
  w_{1,3} & w_{2,3} & -1 & w_{4,3} & ... & w_{p-1,3} & w_{p,3}  \\
  w_{1,4} & w_{2,4} & w_{3,4} & -1 & ... & w_{p-1,4} & w_{p,4}  \\
  ... ...  \\
  w_{1,p} & w_{2,p} & w_{3,p} & w_{4,p} & ... & w_{p-1,p} & -1  \\
\end{pmatrix}
\]
\caption{The Coefficient $\mathcal{C}$ for our general graph Structure}
\end{figure}

The concrete process for the function of Coefficient can also be found in the Forward Process in section 3.3.

In our framework, we have enhanced the representational capacity of each neuron, thereby unlocking the full potential of Neural Networks (NNs). Concurrently, we have eliminated the structural bias that is typically inherent in predefined structures such as traditional tree-like structures or Directed Acyclic Graphs (DAGs). This innovation allows our framework to be more adaptable and less constrained by the limitations of fixed architectural biases, leading to a more flexible and effective NN design.

For large coefficient matrices $\mathcal{C}$, solving the system of equations can indeed be challenging. Our framework addresses this by proposing NM regularization, which serves as an approximation method for $\mathcal{C}$ and incorporates parallel computation to enhance efficiency.

\section{The Connection with DEQ}
Previous research has observed the existence of a stable point in an inﬁnite-level NN structure with identical weights. DEQ addresses this issue by modeling the problem across inﬁnite levels and simplifying it to an implicit function. Specifically, achieving a ﬁxed point is equivalent to solving the root of the implicit function. In our approach, we ﬁnd that our general graph structure can also be managed by solving a system of functions.

In fact, DEQ focuses on infinite levels with the same weight. Suppose treating the infinite path as a circle. The essential research object of DEQ is also a cyclic graph. In this paper, we unveil the fundamental nature of the fixed point, recognizing its role as a solution within our synchronous neural module structure. Our neural module not only assists in finding its essence but also enables concurrent visualization of the implicit function.	

Furthermore, adhering to the Universality of Single-layer DEQ, it is established that multiple implicitly hidden layers are tantamount to a single implicitly hidden layer. As a result, DEQ encounters limitations in managing multiple implicitly hidden layers.

Most importantly, DEQ struggles when managing a larger number of nodes as introduced in the paper. These layers must solve the implicit function. In our paper, by considering implicit hidden layers from infinite levels to elegant general graphs, we would organize the neural units well to optimize the efficiency as well as the performance of the model. To address this, we introduce NM regularization that organizes these nodes into neural modules. In contrast to the DEQ, our framework provides a methodology for handling multiple neural modules. Our analysis of the regularization encompasses both theoretical foundations and experimental parameters, enabling our framework to manage larger graphs effectively. 

Furthermore, the organized independent, balanced neural models achieve superior results on both efﬁciency and performance, as previously demonstrated. These neural modules improve performance by reducing overﬁtting and improving efﬁciency through parallel computation. Previous work like DEQ did not discuss how to organize the neural units to optimize efficiency and performance.

\subsection{The Deeper Understanding with DEQ}

The field of NN structure has also seen the emergence of innovative structures that challenge the status quo. A significant example is the introduction of implicit hidden layers, particularly in DEQ. The developers of DEQ found that by permitting information to propagate through constant weights across an indefinite number of layers, they could achieve a fixed point. This realization enabled them to encapsulate the impact of an infinite layer structure using a system of equations, effectively saving on memory and computational resources.

Nevertheless, the theoretical foundations of these findings remain largely unexplored. There is a noticeable absence of in-depth analysis concerning the significance of the fixed point and the reasons behind why an infinite sequence of unvarying weights would result in such fixed points. This gap in understanding presents an opportunity for further research that could provide deeper insights into the behavior of neural networks with cyclic connections and their potential advantages over traditional architectures. Within these cycles, the neuronal units are interconnected and interdependent. For a cyclic structure, the flow of information recurs through multiple iterations until it achieves an equilibrium state, which is then conveyed to other neuronal units. This equilibrium state is, in essence, the fixed point of the implicit hidden layers.

The propagation of information flow in a cyclic structure leading to a balanced situation or a fixed point can be attributed to the interdependence of neuron units within the network. As outlined in our previous work, the neuron units are not independent but rather rely on each other, effectively forming a complex system of equations that must be satisfied.

In a cyclic structure, the information flow iterates through the network, with each unit's state being influenced by the states of other units within the same neural module. This iterative process continues until the system reaches a point where the changes between iterations become minimal, indicating that the system has converged to an equilibrium state. This equilibrium state is the fixed point where the system's dynamics stabilize, and no further significant changes occur.

The process of reaching this balanced situation in a neural network can be likened to solving a system of equations numerically, such as with the Newton-Raphson method, which is an iterative approach used to find successively better approximations to the roots (or zeroes) of a real-valued function. Similarly, in a cyclic neural network structure, the information flow iterates until it finds a stable solution that satisfies the system's equations, which corresponds to the fixed point.

In summary, the propagation to a balanced situation or fixed point in a cyclic neural network structure is a result of the iterative solution of the system of equations formed by the interdependent neuron units, seeking a stable state that represents the network's equilibrium.

\section{The Connection to Cyclic Structure with Forward-Forward Algorithm}
For Cyclic Structure with Forward-Forward Algorithm [40], the information flow iterates through the network, with each unit's state being influenced by the states of other units within the same cycle. This iterative process continues until the system reaches a point where the changes between iterations become minimal, indicating that the system has converged to an equilibrium state. This equilibrium state is the fixed point where the system's dynamics stabilize, and no further significant changes occur.

The process of reaching this balanced situation in a neural network can be likened to solving a system of equations numerically, such as with the Newton-Raphson method. The Newton-Raphson method is an iterative approach used to find successively better approximations to the roots (or zeroes) of a real-valued function. Similarly, in a cyclic neural network structure, the information flow iterates until it finds a stable solution that satisfies the system's equations, which corresponds to the fixed point.

In summary, the propagation to a balanced situation or fixed point in a cyclic neural network structure is a result of the iterative solution of the system of equations formed by the interdependent neuron units, seeking a stable state that represents the network's equilibrium.

In the key algorithm presented in Cyclic Structure with Forward-Forward Algorithm [40], information propagation through the cyclic structure for a predefined iterations is an attempt to numerically approximate the solution to a system of equations. 

A significant aspect of Cyclic Structure with Forward-Forward Algorithm is its adoption of the forward-forward algorithm, which was introduced by Hinton in 2022 [45]. Additionally, the paper introduces a local loss function based on cross-entropy. This approach is designed to enhance the discriminatory capacity of each individual neuron unit, thereby improving the overall performance of the neural network. The algorithm is structured such that each neuron unit is primarily influenced by its neighboring units. They optimize the model through local learning, employing greedy methods to bypass the traditional backward propagation process, as proposed by Hinton in 2022.

For Cyclic Structure with Forward-Forward Algorithm, the graph structure is predefined, which allows for a simplification of the interrelationships between nodes as needed. This setup is conducive to the forward-forward algorithm, enabling it to operate more efficiently within the defined framework of the neural network's architecture.

On the other hand, in our previous framework, we approached the graph structure search within a complete graph space. Unlike traditional layered concepts, where each neural unit has the potential to connect with any other neuron, in a complete graph structure, every neuron unit is essentially a neighbor to every other neuron. Consequently, any neuron unit is interconnected with all others within the complete graph. However, the initial search space for a complete graph structure can be quite complex. Therefore, for our framework, we aim to propose a method to search for an optimized structure within the complete graph search space, thereby enhancing our model's performance in terms of both performance and efficiency.

To summarize, Cyclic Structure with Forward-Forward Algorithm encompasses three primary points:

A. Leveraging a predefined graph structure, the paper integrates cross-entropy as a local loss function for each Neural Unit and its neighboring neurons, aligning with the forward-forward algorithm proposed by Hinton in 2022. This strategy facilitates ensemble learning with local optimization techniques akin to greedy methods within the information flow.

B. The method addresses the cyclic structure by setting a predefined number of propagation times, which essentially approximates the solution to the system of equations that captures the interdependencies within the graph structure. This is achieved through a predefined number of iterations using numerical methods, as previously analyzed. This approach cannot ensure deterministic outcomes of the proposed solution under all operational conditions.

C. The cyclic structure in the Forward-Forward Algorithm does not account for the size of the cycle, which plays a critical role in determining the framework's complexity. This complexity directly impacts both the performance and efficiency of the algorithm.

To provide preliminary validation, we conducted comparative experiments on the Facebook Dataset by converting the multi-class classification task to a binary formulation. As shown in Table 2, this modified benchmarking framework demonstrates higher accuracy compared to the Cyclic Architecture with Forward-Forward Algorithm.
\begin{table}[!htbp]
	\centering
     \small
	\caption{The error of algorithms}
	\begin{tabular}{lllllll} 
		\toprule
		\cline{1-7}  
            &    & 60 & 80 & 100 & 200 & 300 \\
            & CFF & 0.0496 & 0.0512 & 0.0532 & 0.0522  & 0.0587 \\
	       & NMs\&NM & \textbf{0.0317} & \textbf{0.0322} & \textbf{0.0366} & \textbf{0.0398} & \textbf{0.0412} \\
	\bottomrule
\end{tabular}
\label{tbl:table1}
\end{table}

\section{The Connection with OPTNET}
In earlier research, OPTNET examines the interconnections among nodes within the same level in traditional tree-like structures. Their approach involves implementing Quadratic Programming within these nodes, which introduces signiﬁcant bias. Furthermore, OptNet uses parameters derived from quadratic problems for backpropagation, determined by nodes from the previous layer which brings difficulties to regularization as illustrated before. The defects of OPTNET make it challenging to ﬁne-tune its performance as well as overall efficiency.


In contrast, our neural module is nonlinear and exhibits sufﬁcient ﬂexibility to accommodate any compact function without introducing bias, as afﬁrmed by the Universal Neural Module Approximation Theorem. Additionally, we can readily control the complexity of the parameters to enhance performance and efficiency simultaneously.

\section{Algorithm}
To summarize, we conclude our framework in Algorithm 1, Algorithm 2, and Algorithm 3.

\begin{algorithm}[!htbp]
\caption{Propagation}
\begin{algorithmic}[1] 
\State \textbf{Input:} 
    \( X^{in} \); \( \mathcal{C} \); Input layer weights $W^1$; Output layer weights $W^m$; Propagation type $PT$;
\State \textbf{Output:} 
   The updated vector \( X \) for the nodes.

\If{$PT$ == Forward}
   \State Initialize $X = X^{in} \cdot W^{1T}$.
\EndIf

\If{$PT$ == Backward}
   \State Initialize $X = X^{in} \cdot W^{m}$.
\EndIf

\State Calculate Neural Modules \{NMs\} by Tarjan algorithm on \( \mathcal{C} \).
\State Initialize $Node\_Queue=\{n_i\}$ with $InArc(n_i)=0$.
\State Initialize $NM\_Queue=\{NM_i\}$ with $InArc(NM_i)=0$.
\While $Node\_Queue$ is not null and $NM\_Queue$ is not null
    \For{\( n_i \) in \( Node\_Queue \)}
       \If{$PT$ == Forward}
	       \State Forward process as original NN.
        \EndIf
        \If{$PT$ == Backward}
           \State Backward process as original NN.
        \EndIf
        \State Delete edges of $OurArc(n_i)$ in  \( \mathcal{C} \) .
    \EndFor

    \If{$PT$ == Forward}
	    \State Solve equations 1,9 for \(NM_i \in NM\_Queue\) in parallel.
    \EndIf
    \If{$PT$ == Backward}
        \State Solve equations 2 for \(NM_i \in NM\_Queue\) in parallel.
    \EndIf
    \State Delete edges in $NM\_Queue$.

    \For{\( n_i \)}
        \If{$InArc(n_i)$ == 0 and $OutArc(n_i)$ != 0}
            \State Enqueue $Node\_Queue$ by $n_i$.
        \EndIf
    \EndFor

    \For{\( NM_i \)}
        \If{$InArc(NM_i)$ == 0 and $NM_i$ has edges}
            \State Enqueue $NM\_Queue$ by $NM_i$.
        \EndIf
    \EndFor

\EndWhile
\If{$PT$ == Forward}
    \State  \(X = f(X) \)
\EndIf
\If{$PT$ == Backward}
    \State  \(X = f'(X) \)
\EndIf
\State \Return \(X\)
\end{algorithmic}
\end{algorithm}

\begin{algorithm}[!htbp]
\caption{Training Process}
\begin{algorithmic}[1] 
\State \textbf{Input:} \( X^0 \); \( Y \); $\gamma$; $\mathcal{K}$;$k$; Input layer weights $W^1$; Output layer weights $W^m$; 
\State \textbf{Output:} $\mathcal{K}$;$W^m$;$W^1$;$\gamma$
\State Initiate a random matrix $\mathcal{K}$ for the complete graph.
\While{not converged}
\State Approximate $\mathcal{K}$ by parameter $\gamma$ to get $\mathcal{C}$.
\State Run Propagation with \( X^0 \),\( \mathcal{C} \),\( W^1 \),\( W^m \),Forward.
\State Calculate by $\widetilde{Y} = X \cdot W^{mT}$.
\State Run Propagation with \( \nabla Y \),\( \mathcal{C} \),\( W^1 \),\( W^m \),Backward.
\State Update $\mathcal{K}$ by equation 4.  
\State Update $W^m$ by equation 5. 
\State Update $W_1$ by equation 6. 
\State Update $\gamma$ by the $k$th largest absolute value in $\mathcal{K}$. 
\EndWhile

\end{algorithmic}
\end{algorithm}

\begin{algorithm}[!htbp]
\caption{Predicting Process}
\begin{algorithmic}[1] 
\State \textbf{Input:} \( X^0 \); $\gamma$; $\mathcal{K}$; Input layer weights $W^1$; Output layer weights $W^m$ ;
\State \textbf{Output:} Predicted $\widetilde{Y}$.
\State Approximate $\mathcal{K}$ by parameter $\gamma$ to get $\mathcal{C}$.
\State Run Propagation with \( X^0 \),\( \mathcal{C} \),\( W^1 \),\( W^m \),Forward.
\State Calculate by $\widetilde{Y} = X \cdot W^{mT}$.
\State \Return $\widetilde{Y}$
\end{algorithmic}
\end{algorithm}

\section{The supplementary materials for the experiments}

In this section, we provide a detailed description of the experimental setup, which is presented in Table 3. 

Additionally, as evidenced in this section, the parameters within our framework are adjustable, allowing for optimization and fine-tuning to achieve better performance. In the context of larger neural modules, these critical edges may become less frequent as the process of NM regularization unfolds. This approach encourages the model to concentrate on the most significant connections, thereby improving the efficiency and performance of neural modules.

\begin{table*}[!htbp]
	\centering
     \small
	\caption{Experimental Environment}
	\begin{tabular}{ll} 
		\toprule
	CPU & Gen Intel(R) Core(TM) i9-12900H   2.90 GHz\\
	Cores & 16\\
	Memory & 32G\\
     GPU & NVidia GeForce RTX 3060\\
     Graphics Memory & 12G\\
     Other Parameters & Refer to Figure 16\\
	\bottomrule
\end{tabular}
\end{table*}

\subsection{Performance for Gas Dataset}

We now report the comprehensive benchmarking results of our proposed framework on the Gas Dataset, shown as Table 4 and Figure 6, which was omitted in the main manuscript due to space constraints.

\begin{table}[!htbp]
	\centering
     \small
	\caption{The error of algorithms for Gases Concentration Dataset}
	\begin{tabular}{lllllll} 
		\toprule
		\cline{1-7}  
            &    & 60 & 80 & 100 & 200 & 300 \\
            & NN & 0.1293 & 0.1113 & 0.1414 & 0.1639  & 0.1391 \\
	       & DEQ & 0.1233 & 0.1323 & 0.1098 & 0.1278  & 0.1474 \\
	       & DAG &  0.1835 & 0.1248 & 0.1263 & 0.2857  & 0.1293 \\
	       & OPTNET &  0.1774 & 0.1248 & 0.1549 & 0.1113  & 0.1541 \\
	       & NMs & 0.0789 & 0.0810 & 0.0822 & 0.0832 & 0.0856 \\
	       & NMs\&L1 & 0.0756 & 0.0744 & 0.0720 & 0.0701  & 0.0717 \\
	       & NMs\&NM & \textbf{0.0733} & \textbf{0.0710} & \textbf{0.0632} & \textbf{0.0610}  & \textbf{0.0666} \\
	\bottomrule
\end{tabular}
\label{tbl:table1}
\end{table}

\begin{figure}[!htbp]
     \centering
     \begin{subfigure}[b]{0.5\textwidth}
	 \includegraphics[width=\textwidth]{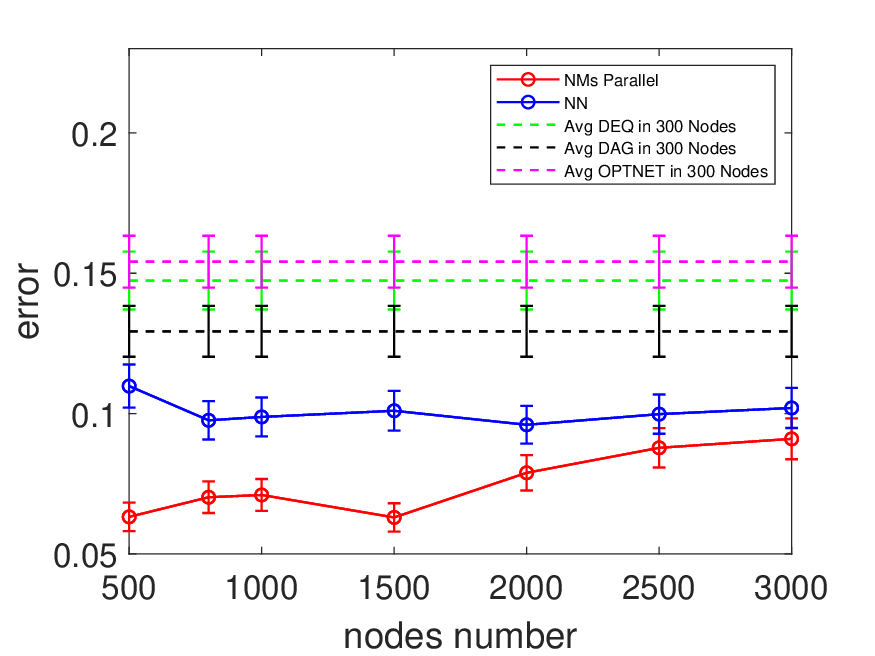}
     \end{subfigure}

	\caption{The performance of NM regularization on up to 3000 nodes for the gas dataset.}
`\end{figure}

\subsection{The Optimization of the Neural Modules}
In this section, we explore the process of optimizing the structure of Neural Modules using NM regularization. As discussed earlier, the results obtained have shown that NM regularization would yield better results. Here, we investigate the parameter-setting strategies for our NM regularization.

\begin{figure}[!htbp]
     \centering
     \begin{subfigure}[b]{0.4\textwidth}
     \includegraphics[width=\textwidth]{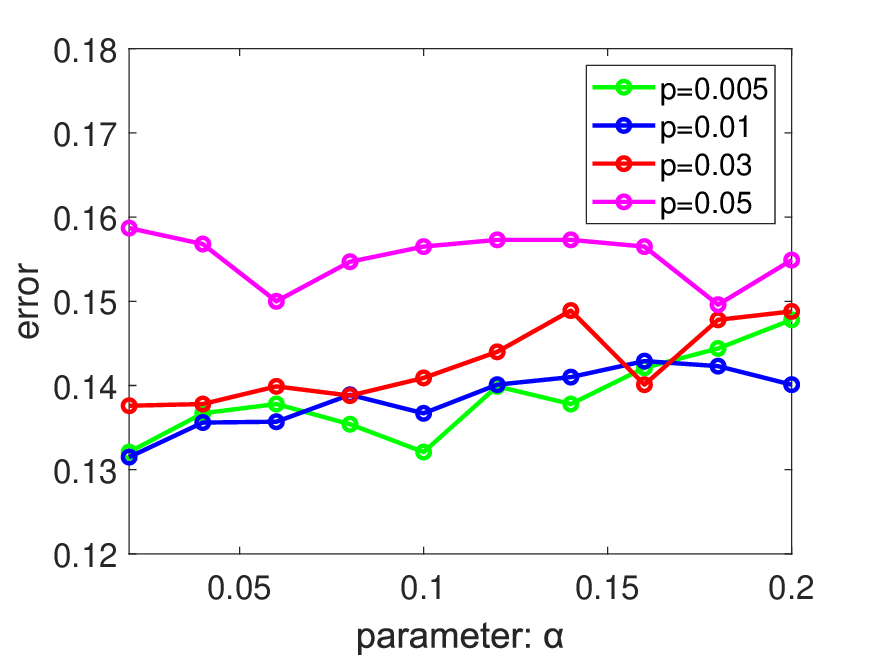}
     \caption{Codon Usage Dataset}
     \end{subfigure}
     \begin{subfigure}[b]{0.4\textwidth}
     \includegraphics[width=\textwidth]{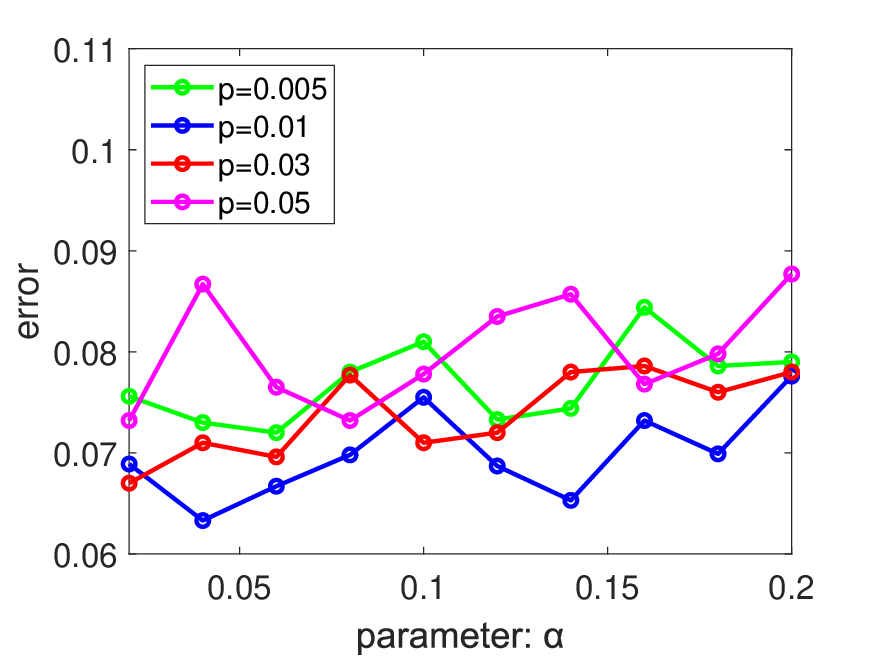}
     \caption{Gas Dataset}
     \end{subfigure}
     \begin{subfigure}[b]{0.4\textwidth}
     \includegraphics[width=\textwidth]{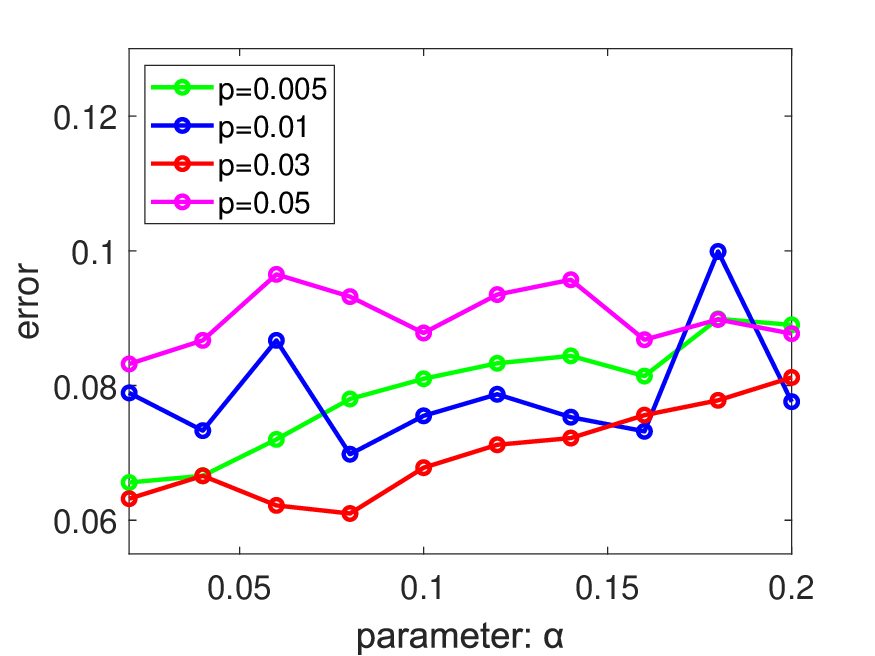}
     \caption{Postural Dataset}
     \end{subfigure}
     \begin{subfigure}[b]{0.4\textwidth}
     \includegraphics[width=\textwidth]{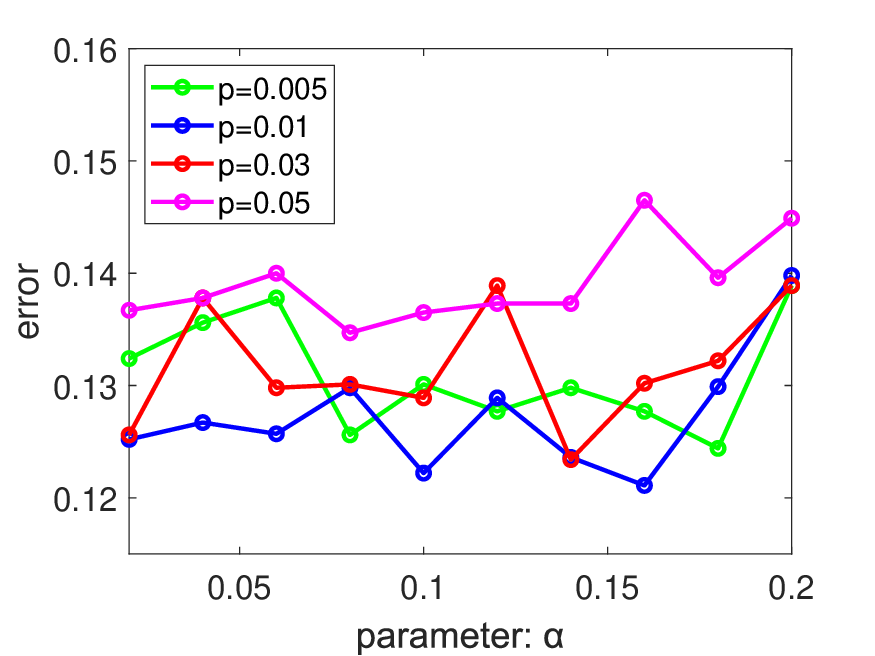}
     \caption{Facebook Dataset}
     \end{subfigure}
     \begin{subfigure}[b]{0.4\textwidth}
     \includegraphics[width=\textwidth]{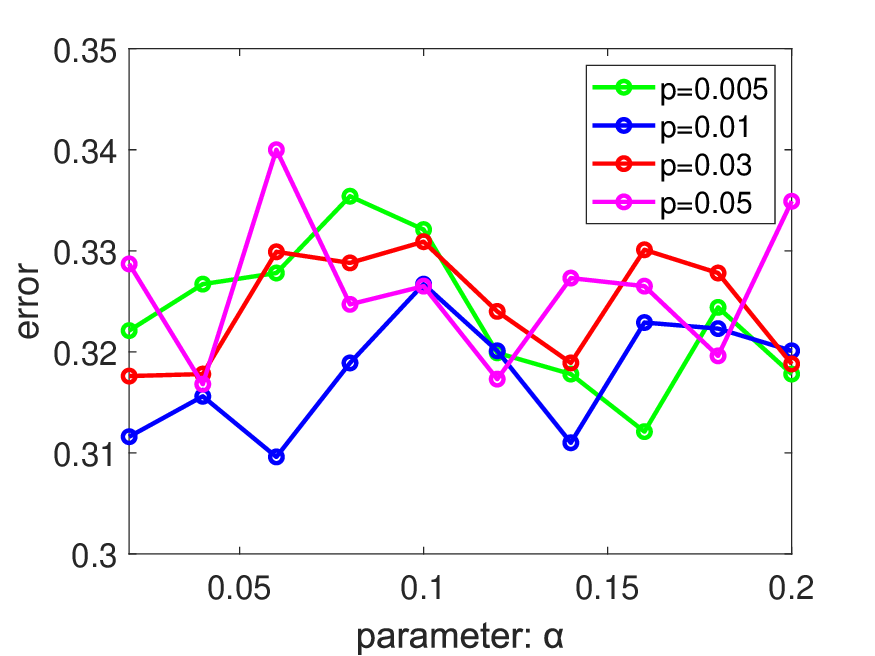}
     \caption{Activity Dataset}
     \end{subfigure}
     \caption{Parameter setting for $NM$ regularization.}
\end{figure} 

The images in Figure 7 illustrate the impact of different parameters on the error rates of our NM regulation across various percentages of approximations for all the Datasets. The tuning process can be seen as an example of our framework. We found that the performance can be optimized by fine-tuning the NM regularization(controlled by parameter $\alpha$) and the percentages of approximations(controlled by parameter $p$). Higher approximation percentages and regularization parameters produce more Neural Modules, which help reduce overfitting problems and in turn, enhance performance as well as robustness. However, overly sparse NMs could introduce extra bias and degrade performance. After standardization, the NM regularization parameter usually performs best near $0.05$ and the approximation on top $0.01$ edges by controlling the parameter $\gamma$. 
 

\subsection{The Effect of NM Regularization}
From the experiments, NM regularization can better capture the trend of the weights in each iteration. It enables accurate weight regularization and formulates effectively balanced neural modules. More importantly, these independent, balanced neural modules bring additional benefits for efficiency. As previously mentioned, the overall efficiency is determined by the size of the largest neural module, especially when they are processed in parallel.

To analyze the effect of NM regularization, consider the probability of the edge’s weight being lower than $\gamma$ denoted as $\theta$ in the current iteration. For a neural module with $q$ nodes, the probability of a new edge integrating into the neural module would be denoted as \begin{equation} 1-\theta^q \end{equation}. In the case of a small $q$ value, NM regularization seeks to increase the probability. Conversely, for larger $q$ values, NM regularization works to decrease the probability. Thus, NM regularization endeavors to minimize the disparity in the probability of a new edge integrating into the neural modules with varying sizes. This effect is illustrated in image A of Figure 8.

\begin{figure}[!htbp]
     \centering
     \begin{subfigure}[b]{0.4\textwidth}
     \includegraphics[width=\textwidth]{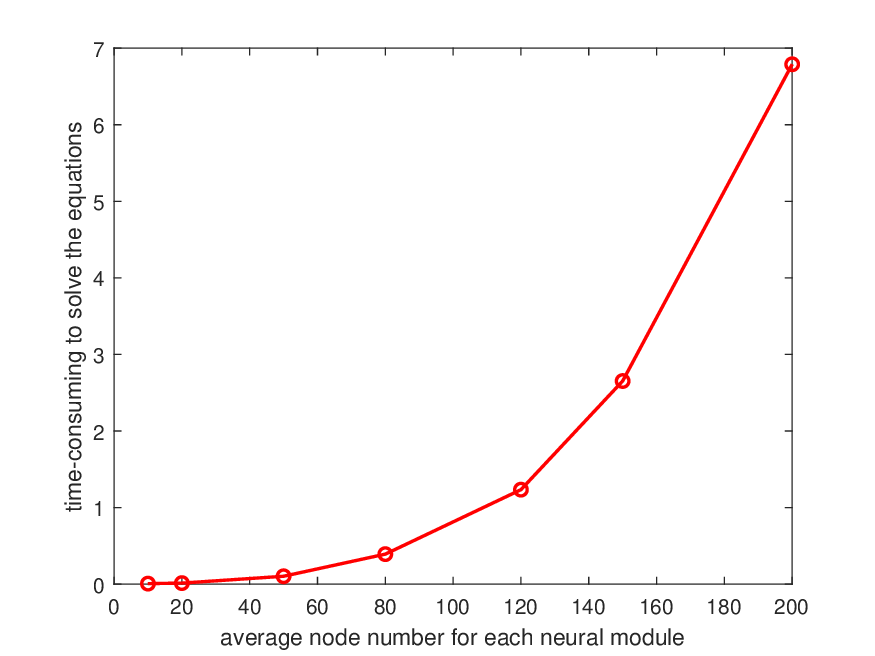}
     \end{subfigure}
        \caption{Image up shows the analysis of NM Regularization and image down displays the time-consumption for different sizes of neural modules which indicates parallel computing's advantage.}
\end{figure}  

Image A in Figure 8 illustrates the analysis of NM Regularization and how balanced Neural Modules are formed through this process. 

Furthermore, Image B in Figure 8 shows the time consumption associated with different sizes of neural modules. As previously mentioned, the neural modules generated are independent and can be calculated in parallel. For instance, if a neural module comprises 50 nodes, the time to solve the system of equations required would be reduced to just 10ms in our experiments. Thus, NM regularization enables sufficient management of larger node counts.

\begin{figure}[!htbp]
     \centering
     \begin{subfigure}[b]{0.4\textwidth}
	 \includegraphics[width=\textwidth]{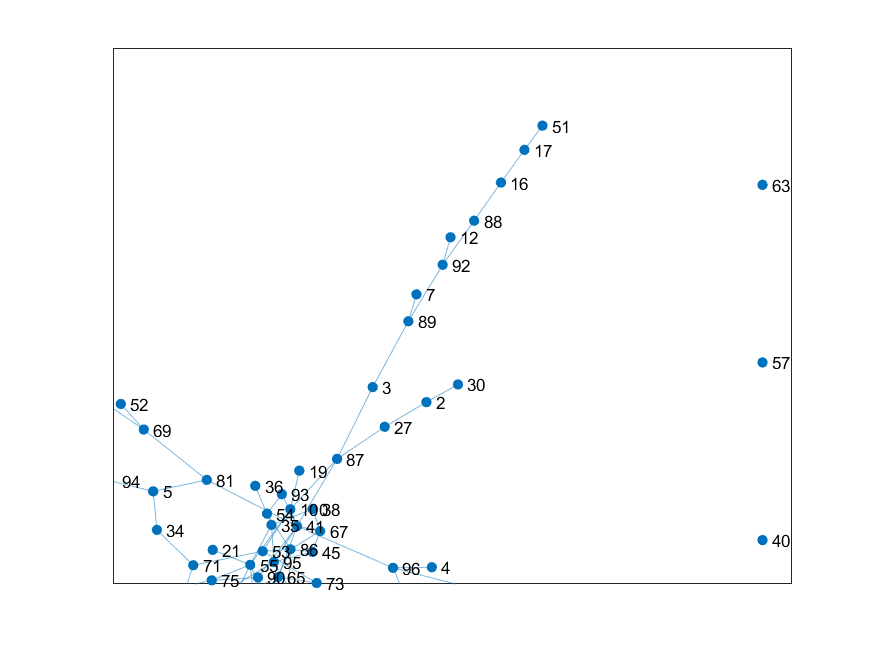}
     \caption{Codon Usage Dataset}
     \end{subfigure}
     \begin{subfigure}[b]{0.4\textwidth}
     \includegraphics[width=\textwidth]{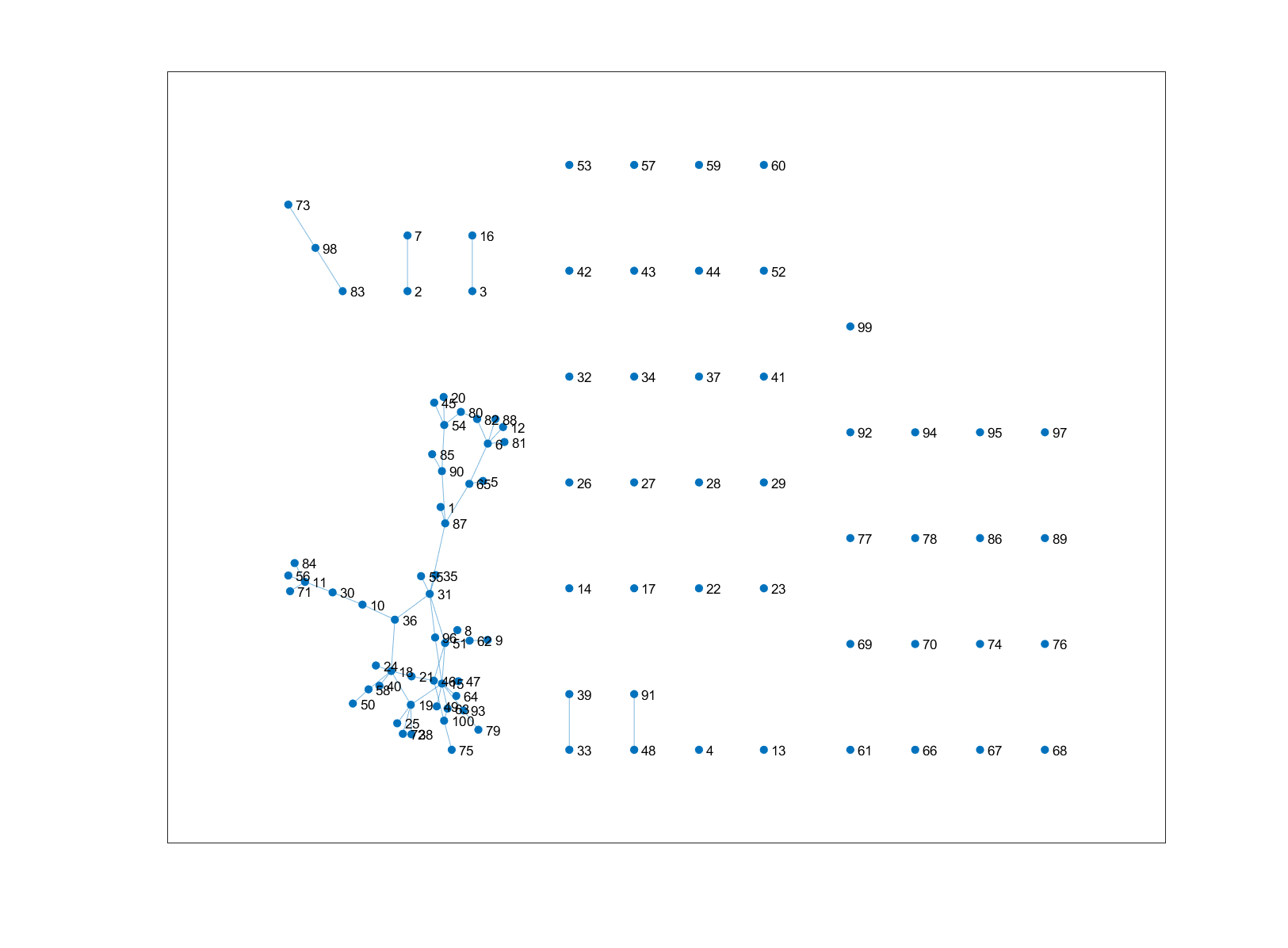}
     \caption{Gas Dataset}
     \end{subfigure}
     \begin{subfigure}[b]{0.4\textwidth}
     \includegraphics[width=\textwidth]{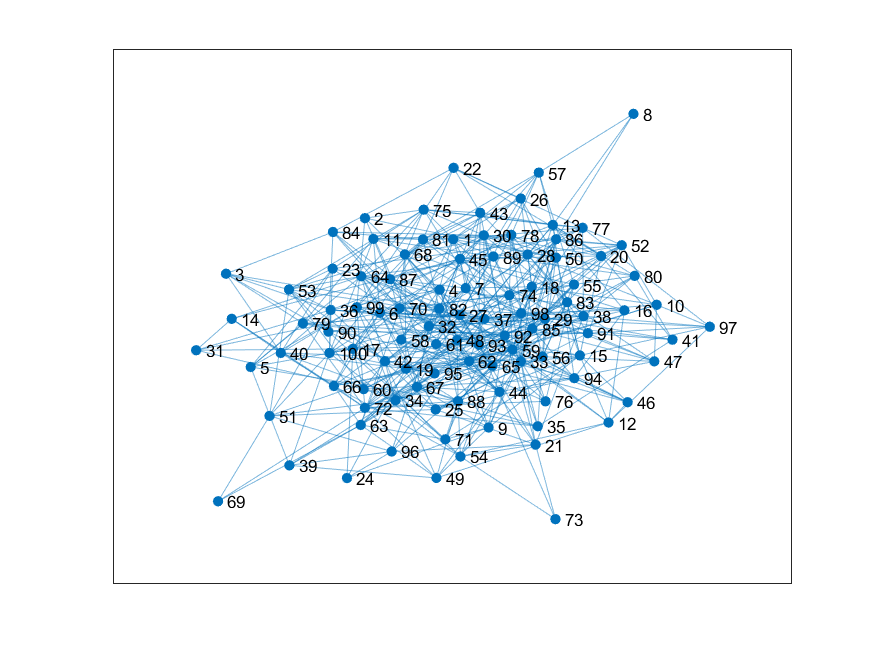}
     \caption{Postural Dataset}
     \end{subfigure}
     \begin{subfigure}[b]{0.4\textwidth}
     \includegraphics[width=\textwidth]{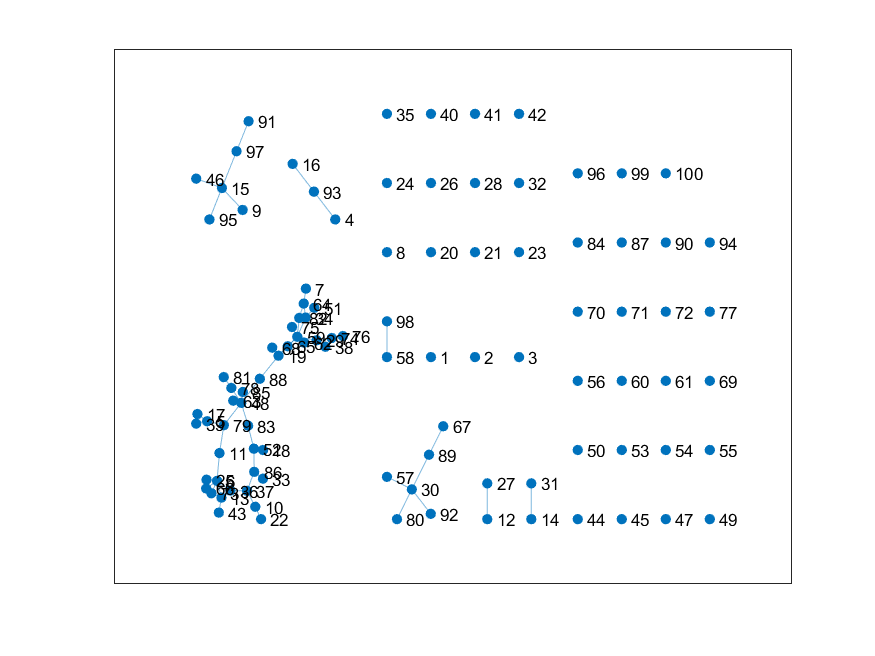}
     \caption{Facebook Dataset}
     \end{subfigure}
     \begin{subfigure}[b]{0.4\textwidth}
     \includegraphics[width=\textwidth]{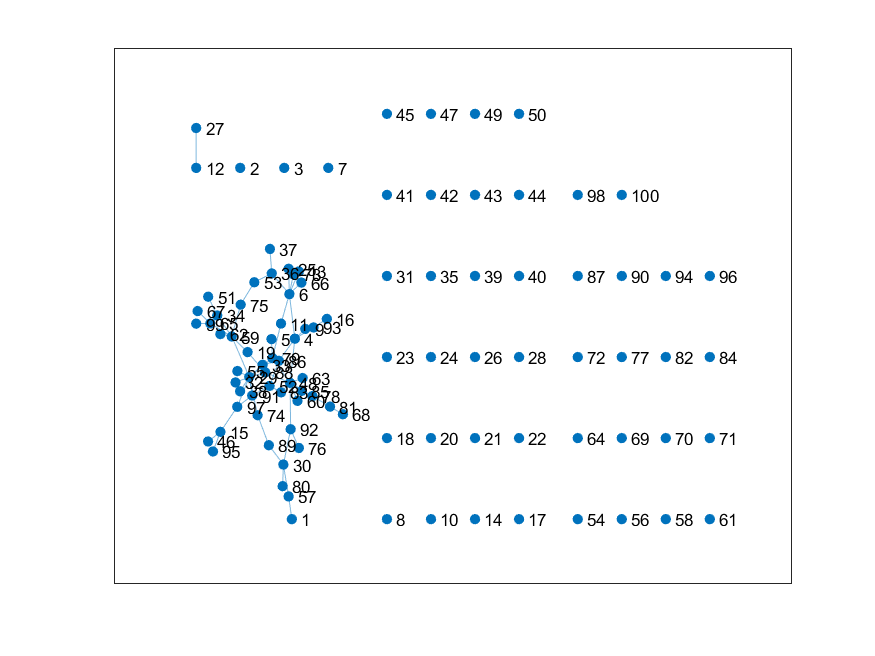}
     \caption{Activity Dataset}
     \end{subfigure}
	\caption{The automatic formed structures by our NM regularization.}
`\end{figure} 



At last, Figures 10,11, and 12 present examples of neural modules generated by our NM regularization for the Datasets, where each black square denotes a neural module. The image on the left shows the neural modules generated at the first iteration, where the effects of NM regulation are not yet apparent. The image on the right, at the 10000th iteration, illustrates how NM regulation achieves independence and balance within the generated neural modules. NM regularization helps break large neural modules into smaller, more manageable ones. This segmentation leads to improved performance and efﬁciency, as previously introduced. 

\begin{figure}[h]
     \centering
	\includegraphics[width=5.6cm]{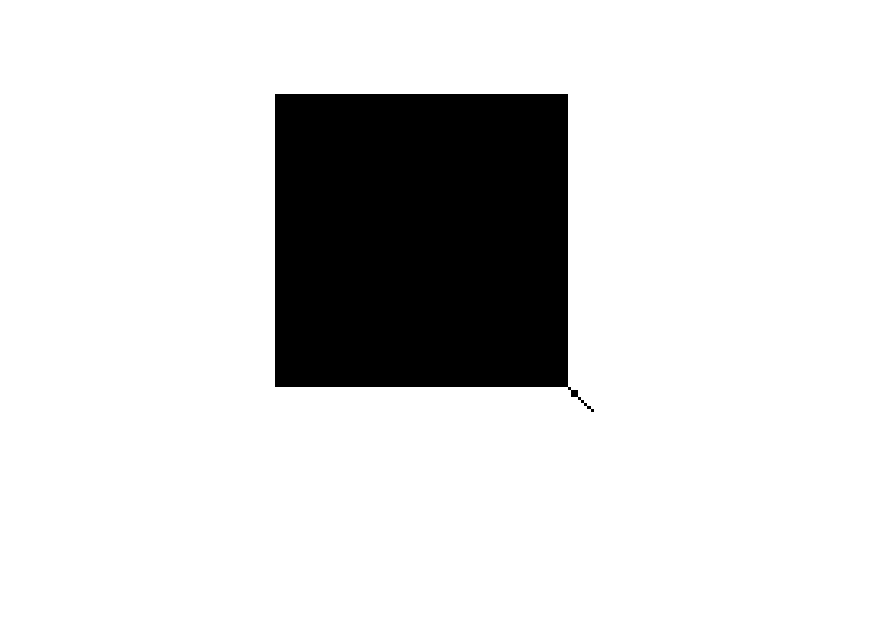}
     \includegraphics[width=5.6cm]{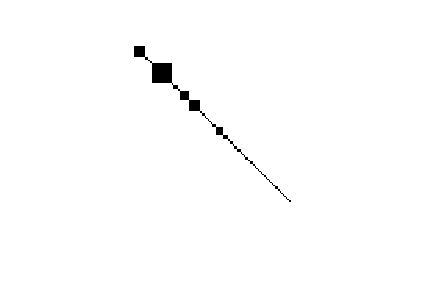}

     \caption{Neural modules generated by our NM regularization for the Codon Usage Dataset with each black square denoting a neural module. The left image depicts the neural modules at the first iteration, and the right image shows the neural modules at the 10000th iteration. NM regularization helps break large neural modules into smaller, more manageable neural modules.}
\end{figure}

\begin{figure}[h]
     \centering
	\includegraphics[width=5.6cm]{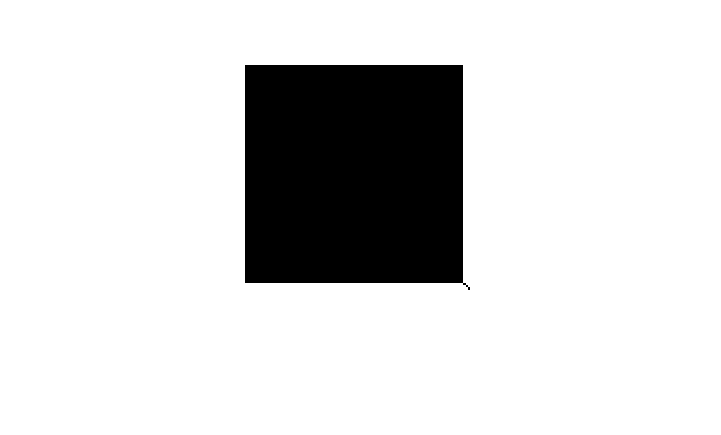}
     \includegraphics[width=5.6cm]{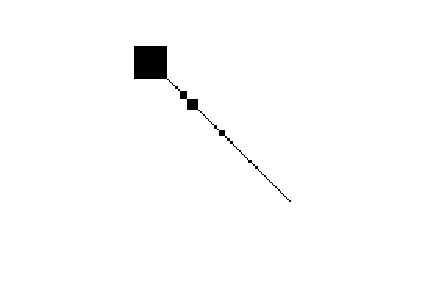}

     \caption{Neural modules generated by our NM regularization for the Gases Concentration Dataset with each black square denoting a neural module. The left image depicts the neural modules at the first iteration, and the right image shows the neural modules at the 10000th iteration. NM regularization helps break large neural modules into smaller, more manageable neural modules.}
\end{figure}

\begin{figure}[h]
     \centering
	\includegraphics[width=6.6cm]{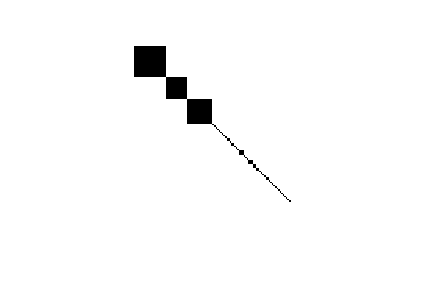}
     \includegraphics[width=6.6cm]{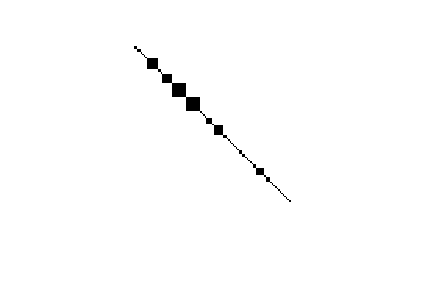}

     \caption{Neural modules generated by our NM regularization for the Postural Transitions Dataset with each black square denoting a neural module. The left image depicts the neural modules at the first iteration, and the right image shows the neural modules at the 10000th iteration.NM regularization helps break large neural modules into smaller, more manageable neural modules.}
\end{figure}

Figure 9 quantitatively compares the topological connectivity patterns generated by our method across three benchmark datasets with 100-node configurations. By analyzing the upper triangular adjacency matrices for simplicity, we observe our framework consistently produces sparse structures matching biological neural network connectivity.

\section{The Proof of Theorem 1}
Note that the array of equations involves an activation function. Therefore, in our model, the system of functions is non-linear and possesses sufficient flexibility to fit any appropriate function, according to our Universal Neural Module Approximation Theorem.

\begin{proof}
In the forward process, the system of equations constructs implicit functions across all nodes within the neural module. Subsequently, any node $n_i$ can be transformed into an explicit function. 
According to the Universal Approximation Theorem [43] [44],  which states that for any continuous function on a compact set, there exists a one-hidden-layer feed-forward network capable of approximating the function, our system satisfies this condition. This is because our array of equations involves an activation function, fulfilling the requirement of the theorem. Therefore, the proof is complete. 
\end{proof}

\section{The Proof of Theorem 2}

\begin{proof}
Consider that \begin{equation} \frac{\partial J_{NM}}{\partial w_{ij}} = \frac{\partial J}{\partial w_{ij}} + \alpha r_{ij} w_{ij}, \end{equation}

Thus \begin{equation} w_{ij} \gets w_{ij} - \eta(\frac{\partial J}{\partial w_{ij}} + \alpha r_{ij} w_{ij}) \end{equation}
\end{proof}

\section{The Proof of Theorem 3}
\begin{proof}
Assuming the $i$th colemn of $\mathcal{C}$ follows a multivariate normal distribution: \begin{equation} w_{:i} \sim \mathcal{N}(0,\tau^2 diag(r_{:i})^{-1}), \end{equation} Consider the Bayesian linear regression model $ y = Xw_{:i} + \epsilon$ and $\epsilon \sim \mathcal{N}(0,\sigma^2 I)$. The posterior distribution of the weight $w$ is given by:: 

\begin{align*}
    p(w_{:i}|y,X) &\propto p(y|X,w_{:i})p(w_{:i})\\
     &\propto exp(-\frac{\|y-Xw_{:i}\|^2}{2\sigma^2})exp(-\frac{\|w_{:i}\|^T diag(r_{:i}) \|w_{:i}\|}{2\tau^2})\\
\end{align*}

The maximum likelihood estimate (MLE) of the parameter is obtained by solving the convex optimization problem\begin{equation} \argmin_{w_{:i}} \{\frac{1}{2\delta^2}\|y-Xw_{:i}\|^2 + \sum {\frac{r_{ji}}{2\tau^2}\|w_{ji}\|^2\}}, \end{equation} which equivalent to $NM$ regularization as \begin{equation} \argmin_{w_{:i}} \{\|y-Xw_{:i}\|^2 + \alpha \sum{r_{ji}\|w_{ji}\|^2}\}. \end{equation}

As previously presented,
\begin{align*}
    log p(w_{:i}|y,X) &\propto -\frac{1}{2\delta^2}\|y-Xw_{:i}\|^2-\frac{1}{2 \tau^2}w_{:i}^T diag(r_{:i}) w_{:i}\\
    &\propto -\frac{1}{2\delta^2}(w_{:i}^T X^T X w_{:i}-2w_{:i}^T X^T y)-\frac{1}{2 \tau^2}w_{:i}^T diag(r_{:i}) w_{:i}\\
    &\propto -\frac{1}{2}w_{:i}^T(\frac{X^T X}{\delta^2}+\frac{1}{\tau^2} diag(r_{:i}))w_{:i}+\frac{w_{:i}^T X^T y}{\delta^2}\\
\end{align*}

In comparison with the standard normal distribution, Theorem 2 yields the posterior distribution.
\end{proof}

\section{The Proof of Theorem 4}

\begin{proof}
As established in Theorem 4, the weights of the parameters adhere to a normal distribution. Consequently, when we prune edges in our framework based on the absolute values of these weights, we observe a folded normal distribution in each iteration. For simplicity, we will now present the cumulative distribution function (CDF) of the standard folded normal distribution: \begin{equation} P(w) = 2\Phi(\frac{w}{\Delta}) - 1. \end{equation} For NM regularization, $\Delta$ of the standard folded normal distribution decreases. The learned parameters exhibit a tendency to approach zero. Consequently, we have \begin{equation} \|w_t-\tilde{w_t}\|^2<\|w_0-\tilde{w_0}\|^2. \end{equation}

Follow Tao Lin in 2020, let $\gamma = \frac{c}{\sqrt{T}}$, $c=\sqrt{\frac{f(w_0)-f(w_*)}{LG^2}}$ and $T$ be the number of iteration.
\begin{align*}
    \mathbb{E}\|\nabla u\|^2 &\leq \frac{f(w_0)-f(w_*)}{\gamma(T+1)} + L \gamma G^2 + \frac{L^2}{T+1} \sum_{t=0}^T \mathbb{E} \|w_t-\tilde{w_t}\|^2 \\
     &\leq \frac{2(f(w_0)-f(w_*))}{\gamma(T+1)} + L \gamma G^2 + L^2 \|w_0-\tilde{w_0}\|^2 \\
\end{align*}
\end{proof}

\end{document}